\documentclass[runningheads]{llncs}

\usepackage{eccv}

\usepackage{eccvabbrv}

\usepackage{graphicx}
\usepackage{booktabs}

\usepackage[accsupp]{axessibility}  

\usepackage{hyperref}

\usepackage{orcidlink}

\begin{document}

\title{MeshAvatar: Learning High-quality Triangular Human Avatars from Multi-view Videos} 

\titlerunning{MeshAvatar}

\author{Yushuo Chen\inst{1}\orcidlink{0009-0005-7571-3915} \and
Zerong Zheng\inst{2}\orcidlink{0000-0003-1339-2480} \and
Zhe Li\inst{1}\orcidlink{0000-0003-4703-0875} \and 
Chao Xu\inst{2}\orcidlink{0009-0007-0437-7525} \and
Yebin Liu\inst{1}\orcidlink{0000-0003-3215-0225}}

\authorrunning{Y. Chen et al.}

\institute{Tsinghua University, Beijing, China \and
NNKosmos Technology, Hangzhou, China}

\maketitle

\begin{abstract}
  We present a novel pipeline for learning high-quality triangular human avatars from multi-view videos. Recent methods for avatar learning are typically based on neural radiance fields (NeRF), which is not compatible with traditional graphics pipeline and poses great challenges for operations like editing or synthesizing under different environments. To overcome these limitations, our method represents the avatar with an explicit triangular mesh extracted from an implicit SDF field, complemented by an implicit material field conditioned on given poses. Leveraging this triangular avatar representation, we incorporate physics-based rendering to accurately decompose geometry and texture. To enhance both the geometric and appearance details, we further employ a 2D UNet as the network backbone and introduce pseudo normal ground-truth as additional supervision. Experiments show that our method can learn triangular avatars with high-quality geometry reconstruction and plausible material decomposition, inherently supporting editing, manipulation or relighting operations. The code is available at \url{https://github.com/shad0wta9/meshavatar}.
\end{abstract}

\section{Introduction}

Human avatar creation plays a crucial role in the animation and film industries, enabling the efficient creation of character animations by artists. Unfortunately, constructing a high-fidelity avatar for a specific character is a costly and laborious process~\cite{debevec2012light,guo2019relightables}. Conventional graphics pipelines typically involve scanning, texturing, rigging, and skinning, each requiring substantial human efforts.

\begin{figure}[t]
    \centering
    \includegraphics[width=1.0\linewidth]{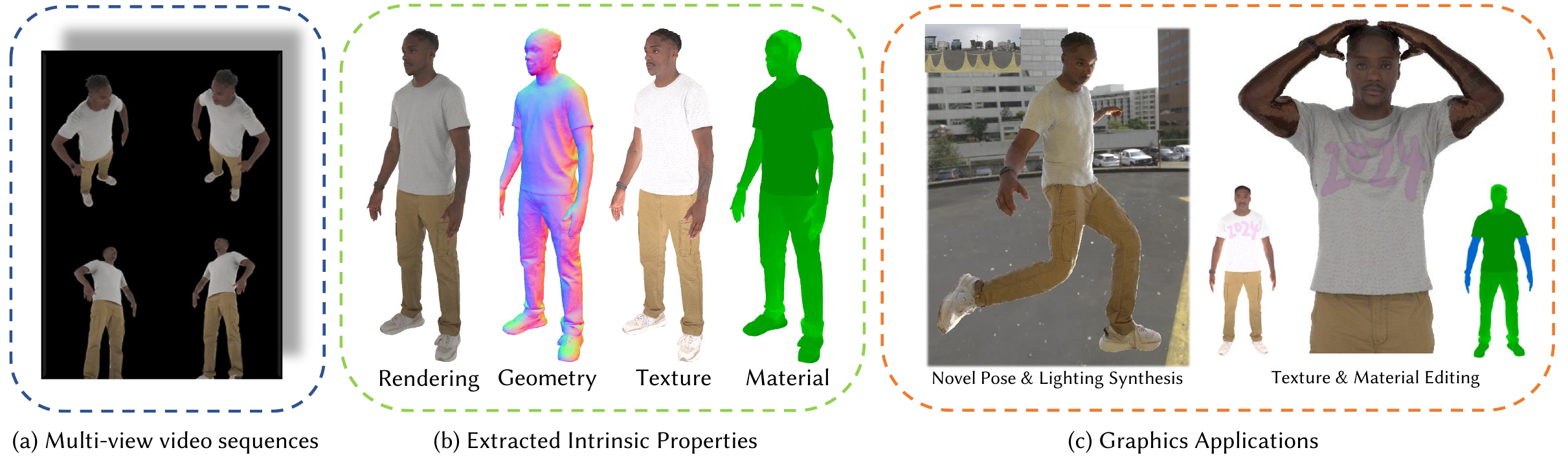}
    \caption{\textbf{Example results by our method.} Given the multi-view videos of a specific subject, our method learns his triangular avatar with the geometry reconstruction and intrinsic material decomposition. After training, the avatar not only supports novel pose synthesis and relighting, but also enables texture editing and material manipulation. 
  In this example, we make the arm metallic and edit the texture of the T-shirt. }
    \label{fig:teaser}
\end{figure}

Benefiting from the huge progress of learning-based 3D modeling, researchers have devoted great efforts in automatically learning 3D human body avatars from images or videos of real humans. 
With the rise of neural radiance fields (NeRF)~\cite{mildenhall2020nerf}, recent works 
tend to represent the 3D character as a pose-conditioned NeRF, and enhance the avatar representation with forward/backward skinning~\cite{deng2020nasa,chen2021snarf,chen2023fast,mihajlovic2021leap,li2022avatarcap,li2023posevocab}, splitting and structuring~\cite{zheng2022structured,zheng2023avatarrex} or UV mapping~\cite{liu2021neural,habermann2023hdhumans}. 
In addition, concurrent works \cite{li2023animatable,zhu2023ash,Zielonka2023Drivable3D,hu2023gauhuman,kocabas2023hugs} introduce 3D Gaussian splatting \cite{kerbl2023gaussian}, a discrete radiance field representation, into human avatar modeling to realize better fidelity and fast rendering.
Although these methods can produce realistic rendering of human avatars, they typically entangle geometry, appearance and lighting into neural networks, 
neglecting the recovery of explicit geoemtric surface 
and limiting many applications in traditional graphics pipelines such as editing and relighting. Although several works attempted to realize inverse rendering for humans in NeRF, they either rely on an inaccurate, physics-agnostic network prediction~\cite{chen2022relighting4d,lin2024relightable}, or require computationally intensive sphere ray tracing~\cite{xu2023relightable}. 

To overcome these limitations, in this paper, we deviate from the current trend and aim to learn human avatars in the form of triangular meshes, which is more compatible with traditional graphics engines and inherently enables physically-based rendering, relighting and editing. 
Although early works has explored mesh-based avatar modeling~\cite{bagautdinov2021driving,xiang2021modeling,xiang2022dressing,habermann2021real,Kwon2023deliffas}, they require per-frame mesh reconstruction \cite{bagautdinov2021driving,xiang2021modeling,xiang2022dressing} or pre-scanned templates \cite{habermann2021real,Kwon2023deliffas}. 
As a result, these avatars cannot be learned in an end-to-end manner and none of them support material decomposition. 

Therefore, we propose a novel hybrid representation for triangular human avatar modeling, enabling complete end-to-end training from multi-view videos. 
Specifically, the learned model consists of an \emph{explicit} triangular mesh geometry (with vertex skinning weights) extracted from an \emph{implicit} SDF field, and an \emph{implicit} dynamic texture and material field conditioned on given poses. 
The explicit mesh and implicit fields are  bridged by differentiable marching tetrahedra (DMTet)~\cite{shen2021dmtet}. 
On one hand, 
mesh-based human geometry can exploit the compatibility with conventional graphics pipelines for editing operations. 
Moreover, hardware-accelerated ray tracing algorithm could be applied for more realistic rendering and better geometry recovery and intrinsic material decomposition~\cite{hasselgren2022shape}. 
On the other hand, the implicit representation delivers higher representation flexibility while inherently regularizing spatial smoothness on both the geometry and the material, which enables effective, end-to-end learning of triangular avatars without any initial guess for the 3D geometry.
After training, the learned avatar is capable of both novel pose synthesis and relighting/editing, as demonstrated in Fig.~\ref{fig:teaser}. 

Nevertheless, acquiring such a representation directly from image observations poses challenges. Dynamic surface reconstruction from sparse views is inherently hindered by the shape-radiance ambiguity, and ensuring model representativeness for dynamic geometric details further complicates the learning process. 
Here we present several technical designs to address these challenges. 
Specifically, we incorporate  shadow-aware Physics-based Rendering (PBR) into the differentiable rasterization process~\cite{hasselgren2022shape}. PBR  accurately models the lighting process and shadow effects by analytically connecting the outgoing radiance with the underlying surface position. 
Compared to implicit radiance fields, PBR can prevent shadow baking, facilitating accurate surface reconstruction. 
To fully harness the benefits of PBR, a geometry reconstruction module capable of recovering high-quality surface details is essential. 
Drawing inspiration from \cite{li2023animatable}, we employ powerful 2D neural networks to encode pose features and model high-frequency geometry information. 
To further enhance geometric details and reduce ambiguities, we leverage normal map prior estimated from image stereo as a weak supervision. Note that this is feasible owing to our triangular mesh representation that allows direct supervision on surface details, which is always difficult in implicit surface representation. 
Benefiting from these technical designs, our method achieves high-quality reconstruction of the shape of dynamic humans, without reliance on explicit surface tracking. 

As shown in the experiments, our method can learn a triangular avatar with high-quality geometry and plausible materials in an end-to-end manner. Moreover, we achieve state-of-the-art dynamic human reconstruction in terms of both geometry and appearance. In summary, our technical contributions are:
\begin{itemize}
    \item We propose an end-to-end pipeline for learning triangular human avatars from multi-view videos. The learned avatars are represented as triangular meshes, which is compatible with traditional graphics pipeline for editing or physics-based rendering. 
    \item We propose a hybrid human avatar representation for triangular human avatar modeling. It bridges explicit triangular mesh geometry with implicit SDF/material fields, and recovers high-quality details by integrating 2D neural networks. 
    \item We achieve high-quality geometry recovery  by PBR differentiable rendering, which can prevent shadow baking and facilitate both surface reconstruction and material decomposition. 
    \item We propose a stereo algorithm based on human prior for normal estimation. The estimated normal maps can be used as a weak supervision to further improve the geometry quality.
\end{itemize}

\section{Related Work}
\label{sec: related work}

\subsection{Human Avatars using Explicit Representations}
\label{subsec: mesh avatars}

Over the last decade, efforts have been made to achieve drivable avatar modeling. Traditional pipelines typically employ  explicit representations to model the shapes of clothed human bodies. Among them, polygon meshes is one of the most popular representations. For example, 
early approaches first reconstruct a character-specific mesh template and animate it through simulation \cite{stoll2010video,guan2012drape} or retrieval \cite{xu2011video}. 
Some works directly deform a generic human body template (e.g. SMPL \cite{loper2015smpl}) to approximate the shape of clothed humans~\cite{alldieck2018video,zhao2022high,burov2021dynamic,kim2022laplacianfusion}. Unfortunately, the quality of avatars is also constrained by the limitations of their deformed SMPL representation.
In the deep learning era, recent works, leveraging convolutional networks as a powerful learning tool, propose to learn non-rigid deformation and dynamic appearance in the UV space of subject-specific mesh templates~\cite{bagautdinov2021driving,xiang2021modeling,xiang2022dressing,habermann2021real,Kwon2023deliffas}. 
These methods usually require pre-scanning a subject-specific template or sophisticated non-rigid tracking throughout the sequence. In contrast, our hybrid representation eliminates these requirements and autonomously learns an  triangular mesh template in an end-to-end fashion.

Apart from mesh-based representation, point clouds  have gained popularity for their flexibility. These approaches mainly focus on modeling the human geometry~\cite{ma2021scale,ma2021power,lin2022learning,zhang2023closet,zheng2023pointavatar}. 
Recently, 3D Gaussian splatting (3DGS)~\cite{kerbl2023gaussian}, an efficient differentiable point-based rendering method, has demonstrated real-time photo-realistic rendering. By incorporating 3DGS into avatar modeling, concurrent works realize high-fidelity human animation results~\cite{li2023animatable,zhu2023ash,Zielonka2023Drivable3D,hu2023gauhuman,kocabas2023hugs,lin2024layga}. However, the unstructured nature of points poses significant challenges for recovering the underlying geometry and/or its intrinsic material properties of human shapes. In contrast, our method is able to recover the geometric shapes as well as intrinsic properties, enabling physics-based rendering and avatar relighting.

\subsection{Human Avatars using Implicit Representations}
\label{subsec: implicit function avatars}

Implicit representations model scene with coordinate-based MLPs that map continuous spatial coordinates to an implicit field, such as signed distance function (SDF)~\cite{park2019deepsdf}, occupancy~\cite{mescheder2019occupancy}, and radiance  fields (NeRF)~\cite{mildenhall2020nerf}.
Since their debut, it becomes a popular trend to adopt implicit representations in avatar modeling framework. 
In the realm of geometric avatar modeling, numerous approaches leverages pose-conditioned SDF field~\cite{saito2021scanimate,wang2021metaavatar,tiwari2021neural,dong2022pina,ho2023learning} or occupancy fields~\cite{deng2020nasa,chen2021snarf,chen2023fast,mihajlovic2021leap,li2022avatarcap} learned from human scans or depth sequences.
To capture the avatar appearance, recent works leverage NeRF as the underlying representation~\cite{peng2022animatable,su2021a-nerf,zheng2022structured,Feng2022scarf,weng2022humannerf,te2022neural,peng2022selfnerf,guo2023vid2avatar,jiang2023instantavatar,jiang2022neuman,li2022tava,wang2022arah,li2023posevocab,dong2022totalselfscan,shen2023xavatar,zheng2023avatarrex} for its impressive learning capability. 
Some research explores hybrid representations that combine the strengths of both explicit and implicit representations~\cite{liu2021neural,chen2023uv,su2023npc,remelli2022drivable,shao2021doublefield}. 
Despite the rapid progress, these implicit function-based methods necessitates dense sampling and extensive network evaluation along pixel rays, suffering from high computational complexity and making physics-based ray tracing extremely difficult. 
In contrast, our representation directly generates a polygon mesh for image rendering, which allows us to seamlessly integrate physics-based ray tracing methods. This enables high-fidelity material decomposition and reconstruction while mitigating the computational burden associated with ray tracing in implicit representations.

\subsection{Neural Inverse Rendering}

Inverse rendering aims to disentangle intrinsic scene properties including geometry, material and lighting conditions from image observations. This is a fundamental yet extremely ill-posed task. Therefore, previous methods~\cite{bi2020deep, goldman2009shape, lawrence2004efficient, nam2018practical, barron2012shape, legendre2019deeplight, dong2014appearance, li2018learning, Iwase_2023} often resorted to assumptions like known geometry, controlled lighting or other physics priors. Recently, advances in neural implicit representations aid in inverse rendering tasks by simultaneously recovering scene properties without explicit geometry guidance~\cite{boss2021nerd, zhang2021nerfactor, zhang2021physg, boss2021neural, srinivasan2021nerv, jin2023tensoir}.  Despite achieving impressive results, the ambiguity in volume rendering lead to low geometric quality. Surface-based inverse rendering methods, on the contrary, directly optimize the underlying geometry with differentiable rasterization, showing higher accuracy in surface reconstruction~\cite{Munkberg2022nvdiffrec,hasselgren2022shape}. 

This progress have also inspired follow-up research on human performance relighting from sparse multi-view videos. A typical pipeline decomposes human performance into canonical reflectance fields and incorporate human templates for motion modeling~\cite{chen2022relighting4d, iqbal2023rana, sun2023neural}. However, these methods often struggle with inaccurate surface reconstruction due to erroneous motion estimation and lack of geometry supervision. The compromised quality of dymanic geometry hampers the modeling of material properties, resulting in unsatisfactory relighting effects. In contrast, our approach excels in high-fidelity surface reconstruction and material decomposition for complex human avatars, overcoming the limitations of existing methods and delivering superior results. Moreover, our triangular representation are more suitable in conventional graphics engines and applications. Concurrent to us, Wang et al.~\cite{WangCVPR2024IntrinsicAvatar}, Xu et al.~\cite{xu2023relightable} and Lin et al.~\cite{lin2024relightable} improve the previous pipelines of relightable avatar learning by introducing explicit ray tracing~\cite{WangCVPR2024IntrinsicAvatar}, hierarchical distance query~\cite{xu2023relightable} or part-wise light visibility estimation~\cite{lin2024relightable}. These techniques allow for better modeling of the lighting process. However, both of them are still built upon implicit representation, making them computationally expensive when synthesizing novel poses and views. 

Some methods have been proposed to enable relighting of human portraits from low-cost inputs such as single images~\cite{ji2022relight,Lagunas2021humanrelighting,tajima2021RHW,Kanamori2018relightinghumans,Jiang2023nerffacelighting} or videos~\cite{wang2023sunstage}. Despite their convenience, these methods cannot change the viewpoints or the body poses, so a deep discussion of these works is out of the scope of this paper.

\section{Method}

\begin{figure*}[t]
    \centering
    \includegraphics[width=\textwidth]{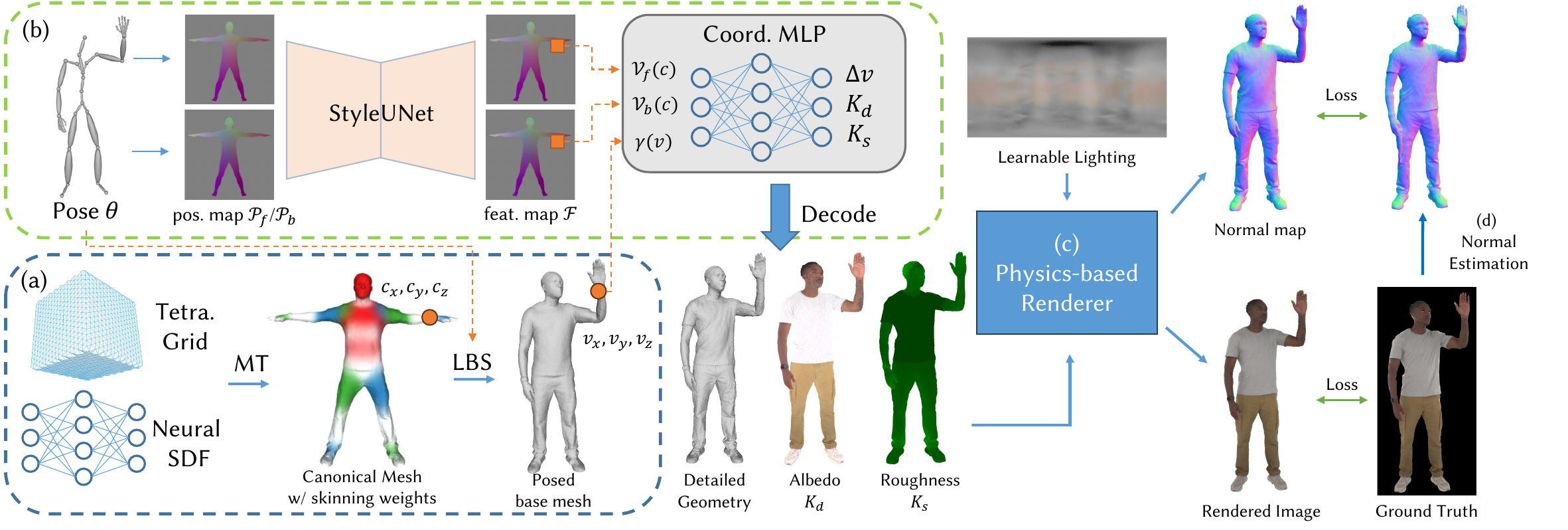}
    \caption{\textbf{Illustration of our method.} Our pipeline learns a hybrid human avatar represented in the form of (a) an explicit skinned mesh and (b) implicit pose-dependent material fields. Such a representation inherently supports (c) physics-based ray tracing and can be trained in an end-to-end manner using (d) normal estimation as an additional supervision signal. }
    \label{fig:method}
\end{figure*}

\subsection{Overview}
Given a collection of multiview videos capturing a clothed human' motions and the corresponding SMPL-X registrations, our task is to build a photo-realistic avatar for this subject in an end-to-end manner, \emph{in the form of an explicit triangular mesh}. 
Such a goal deviates notably from the recently popular trend of NeRF-based avatar modeling, and this departure is particularly significant as it addresses the challenges associated with intrinsic decomposition in human avatar modeling.
However, this task is by no means trivial and poses two critical difficulties. 
One is to construct an accurate mesh template with pose-dependent fine-grained details for clothed human animation, and the other is to address the shape-radiance ambiguity in surface reconstruction. 
To tackle these difficulties, we propose a novel pipeline utilizing a hybrid representation to model triangular clothed human avatars.

As shown in Fig. \ref{fig:method}, our pipeline learns a hybrid human avatar representation, which consists of an explicit skinned mesh extracted from an implicit SDF field using deep marching tetrahedra (Sec.~\ref{sec:method:mesh}), and a pose-dependent detail generation network that produces dynamic vertex offsets (Sec.~\ref{sec:method:details}). 
The intrinsic material properties including albedo and roughness are predicted in the form of volumetric fields, with which our avatar can be rendered using physics-based rendering given the learnable environmental lighting (Sec.~\ref{sec:method:pbr}). All networks are trained in an end-to-end manner supervised by the input images and normal estimations (Sec.~\ref{sec:method:normal}). 


\subsection{Canonical Human Skinned Mesh}
\label{sec:method:mesh}

In this paper, we propose a hybrid representation to model the canonical implicit SDF field of the clothed human. Specifically, we represent the SDF field by a coordinate-based MLP
\begin{equation}
    \mathcal{S}(x): \mathbb{R}^3 \mapsto \mathbb{R}.
\end{equation}
in order to regularize the topology smoothness~\cite{park2019deepsdf}. Instead of performing volume rendering from the SDF field, we leverage marching tetrahedra (MT)~\cite{shen2021dmtet} algorithm to extract the mesh in the canonical space for the following processing. This algorithm differentiably connects the explicit triangular mesh and the implicit SDF field in the canonical space, making it possible to optimize for a variety of shapes in an end-to-end manner. 



To associate the canonical space and the observation (posed) space, a skinning weight field is required for skinning the canonical mesh extracted by DMTet.
Previous works project the spatial points onto the fitted SMPL surface to obtain an initial guess, and learn some residuals for refinements~\cite{peng2021animatable}. 
However, simultaneously solving the geometry and its skinning weights is highly ill-posed. 
Therefore, we follow FITE~\cite{lin2022learning} to spread the skinning weights $\mathcal W^{\mathcal T}$on SMPL surface $\mathcal T$ into 3D space by minimizing the energy:
\begin{align}
    \mathcal{E} =& \lambda_p \int_{p\in \mathcal T} \|\nabla\mathcal W(p) + \nabla_\mathcal T\mathcal W^\mathcal T(p)\|^2 \notag +\\
    & \lambda_\mathcal T\int_{p\in \mathcal T} \|\mathcal W(p) - \mathcal W^\mathcal T(p)\|^2  +  \lambda_r \int_{p\in \mathbb R^3} \|\nabla^2\mathcal W(p)\|^2, 
\end{align}
where $\mathcal W:\mathbb R^3 \to \mathbb R^J$ denotes the diffused skinning weights. After minimizing the above energy, we cache the diffused skinning weights on a $256\times 256\times 256$ tetrahedra grid for subsequent queries. In this way, an arbitrary mesh extracted from the grid could be instantly transformed into any posed space via the linear blend skinning (LBS) algorithm. Specifically, let's denote the vertices on the extracted canonical mesh as $\{\mathbf{c}_n\}_{n=1}^{N_c}$. Given any pose $\mathbf{\theta}\in\mathbb R^{3J}$, the corresponding vertex in the posed space
\begin{equation}
    \mathbf v_n = \sum_{i=1}^J \mathcal W\left(\mathbf{c}_n\right)_i \mathbf{M}_i(\mathbf \theta) \mathbf{c}_n,
\end{equation}
where $\mathbf{M}_i(\mathbf \theta)$ is the rigid transformation for the $i$-th joint.

\subsection{Pose-dependent Geometric Detail Generation}
\label{sec:method:details}


Although the representation in Sec.~\ref{sec:method:mesh} can roughly approximate the shape under different body poses, high-frequency details like the cloth wrinkles are missing due to the low-frequency bias in MLP~\cite{tancik2020fourier}. 
Without high-quality surface geometry, material decomposition and physics-based rendering are impossible. Although previous works like LaplacianFusion~\cite{kim2022laplacianfusion} have represented the high-frequency details with Laplacian coordinates, they require expensive computations at test time. Thanks to our triangular mesh representation, we can represent the geometric details by pure vertex offsets on the skinned mesh, and the problem boils down to accurate vertex offset estimation. 

To this end, we draw inspiration from \cite{li2023animatable} and employ techniques on 2D images to generate high-frequency features. Specifically, consider the vertices on the canonical SMPL template, we assign the posed coordinates to the vertex colors, then acquire the front/back position map $\mathcal{P}_f,\mathcal{P}_b\in \mathbb{R}^{512\times 512\times 3}$ by orthogonally project the colored mesh onto the front/back plane. The position maps are then fed into a UNet $\mathcal{U}$ to obtain the vertex offset maps and an auxiliary feature map, as
\begin{equation}
    \left(\mathcal{V}_f, \mathcal{V}_b, \mathcal{F}\right) = \mathcal{U}\left(\mathrm{concat}(\mathcal{P}_f(\mathbf\theta), \mathcal{P}_b(\mathbf\theta))\right), \label{eq:feat}
\end{equation}
where $\mathcal{V}_f$ and $\mathcal{V}_b$ denotes the front and back vertex offset maps, respectively, and $\mathcal{F}$ is the auxiliary pose feature map. 
Then the high-frequency vertex offsets are directly extracted from the offset map via bilinear interpolation according to the canonical coordinates: 
\begin{equation}
    \Delta \mathbf v_n = \mathrm{1}_{\{\mathbf{c}_{n,z} > 0\}}\mathcal{V}_f(\mathbf{c}_{n,x}, \mathbf{c}_{n,y}) + \mathrm{1}_{\{\mathbf{c}_{n,z} \leq 0\}}\mathcal{V}_b(\mathbf{c}_{n,x}, \mathbf{c}_{n,y}).
\end{equation}
To enhance generalizability of this pose encoding for novel poses, we also apply position map projection onto training space by PCA~\cite{li2023animatable}.

\subsection{Material Maps and Physics-based Rendering}
\label{sec:method:pbr}


Following the conventions in previous works, we use Disney material model~\cite{disney} and assume that the illumination comes from infinity, thus it could be represented as an environmental map. These material parameters and lighting parameters can be optimized through differentiable rendering. Considering inverse rendering itself is a highly ill-posed problem, we manually set the metallic coefficient to zero in order to to regularize the solution complexity for clothed human bodies. 
Although the network in Sec.~\ref{sec:method:details} is powerful enough to generate high-frequency geometric details, tiny errors are still inevitable due to the limited mesh resolution.  
To compensate for this inaccuracy, we make the material map pose-dependent.  
Specifically, for any point $\mathbf{v}$ on the posed human body, we decode its material properties (albedo color, roughness) from the feature map (Eq.~\ref{eq:feat}) by a shallow MLP $\mathcal{M}$:
\begin{equation}
    \left(k_d(\mathbf v), k_s(\mathbf v)\right) = \mathcal{M}\left(\mathrm{concat}(\gamma(\mathbf v), \mathcal F(\mathbf c_x, \mathbf c_y)) \right),
\end{equation}
where $\mathbf{c}$ is the corresponding canonical coordinate calculated via barycentric interpolation, and $\gamma(\cdot)$ is the Fourier positional encoding function~\cite{mildenhall2020nerf}. To guarantee material consistency across poses, the environmental lighting is restricted to be identical for different poses. 

Since the human body is diffusive-dominant and consequently low-frequency environment map is sufficient to recover the materials~\cite{zhang2021physg,Munkberg2022nvdiffrec,hasselgren2022shape}, we apply a differentiable Monte-Carlo (MC) renderer~\cite{hasselgren2022shape} for image rendering. Following the rendering equation~\cite{kajiya1986rendering}, the outgoing radiance $L(\omega_o)$ is computed by the integration of the product of incident radiance $L_i(\omega_i)$ and BRDF function $f_\mathrm{BRDF}(\cdot,\cdot)$ over the upper hemisphere $\Omega$ around the surface normal $\mathbf{n}$, which can be approximated by Monte-Carlo sampling on reflection rays: 
\begin{align}
    L(\omega_o) &= \int_{\Omega} L_i(\omega_i) f_\mathrm{BRDF}(\omega_i, \omega_o) \langle \omega_i, \mathbf{n}\rangle~\mathrm{d}\omega_i \notag \\
    &\approx \frac{1}{N}\sum_{i=1}^N \frac{L_i(\omega_i) f_\mathrm{BRDF}(\omega_i, \omega_o) \langle \omega_i, \mathbf{n}\rangle~\mathrm{d}\omega_i }{p(\omega_i)}.
\end{align}

Note that this sampling process is particularly effective under our assumption of low-frequency lighting. We also apply Multiple Importance Sampling (MIS) and image denoising techniques to further improve the sample efficiency, so that the image can be rendered with only 4 to 8 samples per pixel during training ~\cite{hasselgren2022shape}. 
Moreover, this MC-based renderer is able to account for the occlusions and generate shadows, thereby helping us reconstruct accurate and high-quality materials and create photo-realistic animations. 
Please refer to ~\cite{hasselgren2022shape} for more implementation details about the MC-based renderer. 


\begin{figure*}[t]
  \centering
  \includegraphics[width=\textwidth]{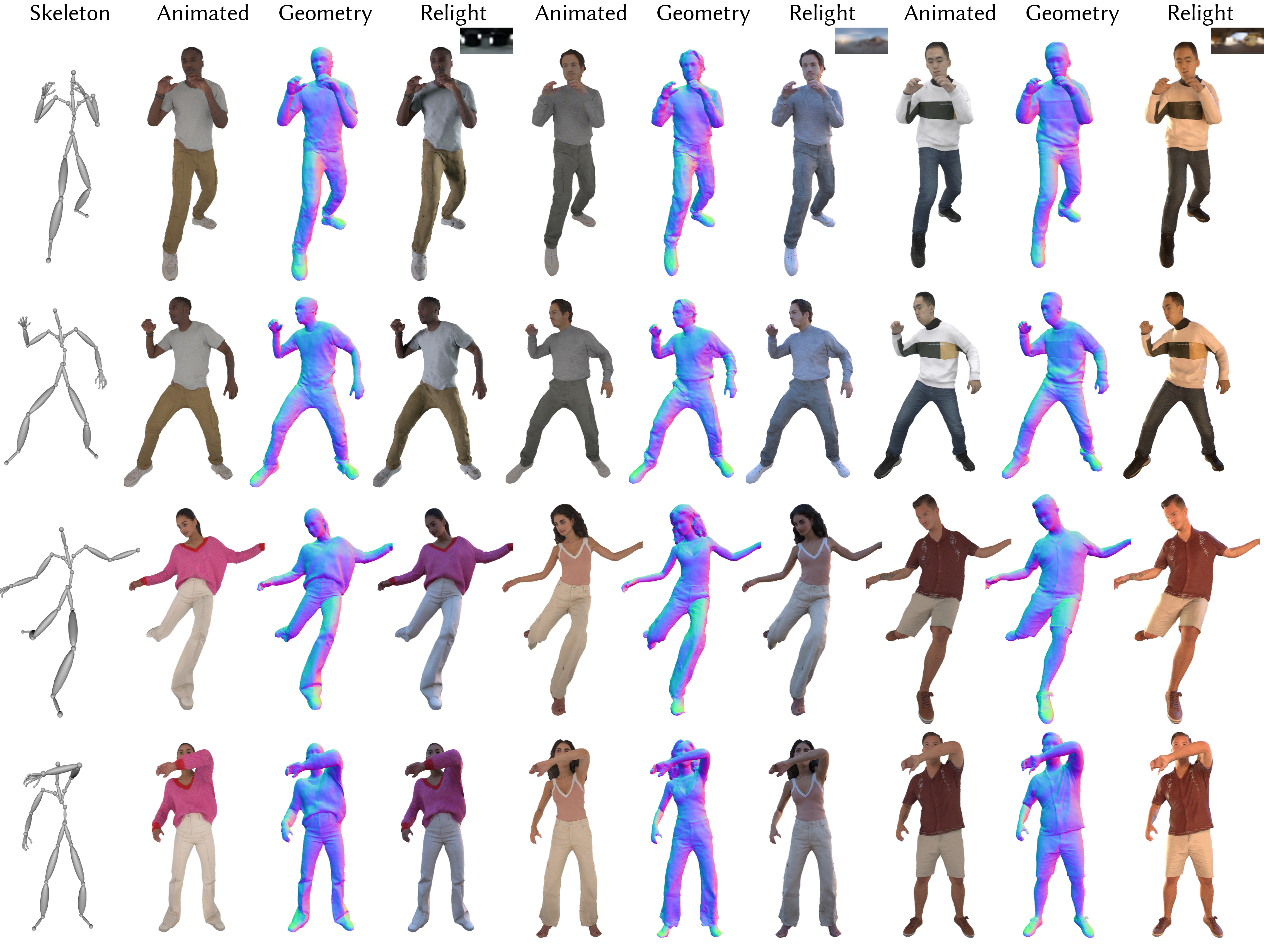}
  \caption{\textbf{Animation and relighting results.} The actors on the same row are driven with the same pose, while the actors on the same column are relighted under the same global illumination.}
  \label{fig:animate}
\end{figure*}

\subsection{Prior-based Normal Supervision}
\label{sec:method:normal}

To further reduce ambiguities and enhance geometric detail reconstruction, we leverage normal maps estimated from the multi-view inputs as an additional supervision signal.
For this purpose, we utilize 3D human scans collected from \cite{yu2021function4d} to train a normal estimation network. 
The network consists of two modules designed to estimate normals in a coarse-to-fine fashion. The coarse module, adapted from RAFT-Stereo~\cite{lipson2021raftstereo}, takes a pair of images from two neighboring views as input and produces a disparity map, as is also done in StereoPIFu~\cite{hong2021stereopifu}. 
The disparity map is then converted into a depth map, from which coarse normals can be analytically calculated. 
After that, a refinement network is employed to further enhance the normal details. This is accomplished by an image-to-image translation module, which refines the coarse normal map by utilizing the RGB image as a condition. Both network modules are trained with ground-truth supervision, which is rendered from the 3D scans in a similar way to PIFuHD~\cite{saito2020pifuhd}. Once trained, the normal estimation network is applicable to different datasets.

To use the normal prior in our avatar learning pipeline, we simply apply it to the multi-view image inputs to obtain the normal estimations. These estimations serve as pseudo ground-truth to construct an additional loss for supervising the avatar geometry.
The detailed learning objective will be presented in the supplemental material.

\begin{figure*}[t]
    \centering
    \includegraphics[width=\linewidth]{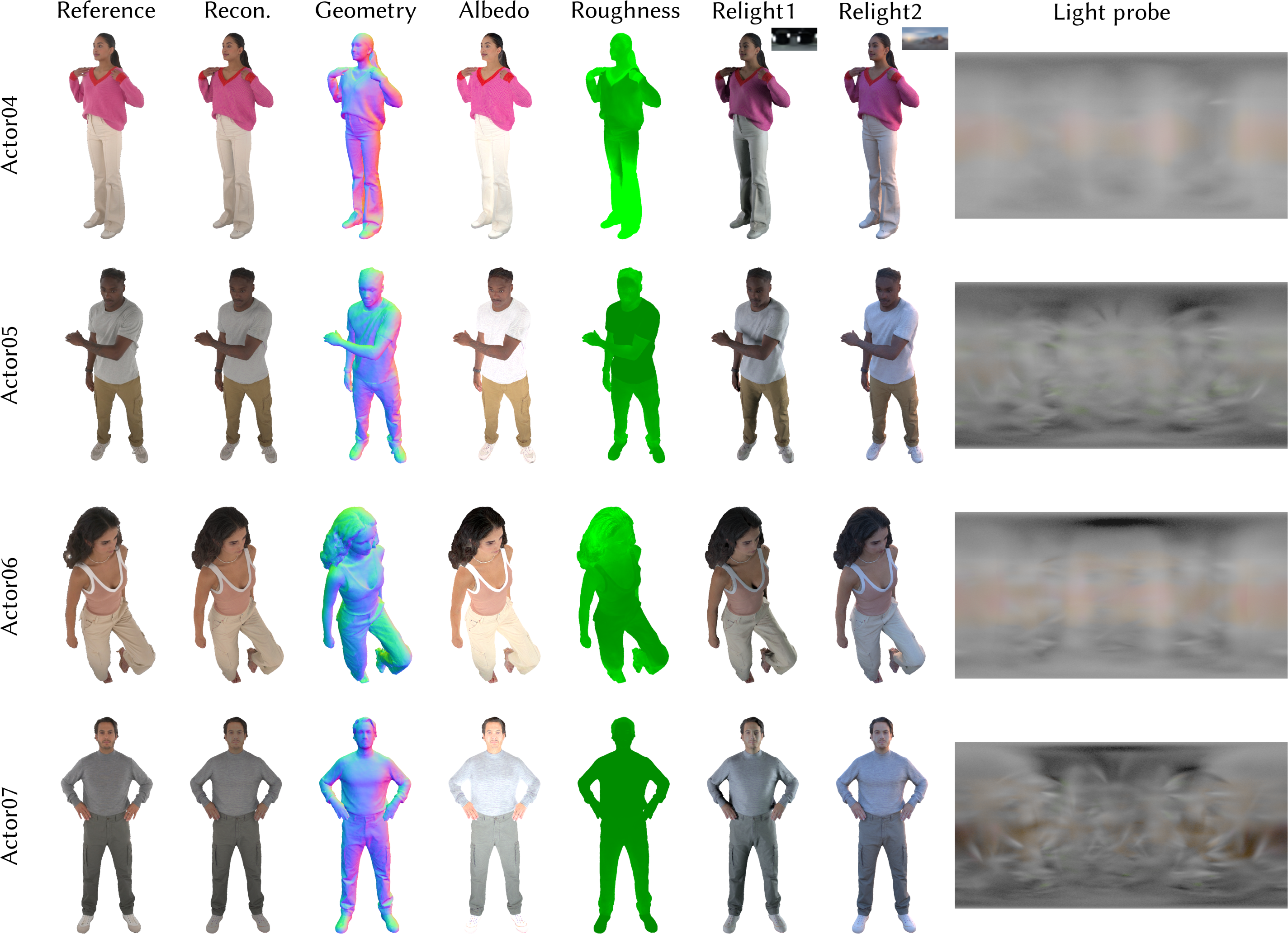}
    \caption{\textbf{More reconstruction results}, including reconstructed geometry, albedo color, material roughness, and light probes. }
    \label{fig:rec-more}
\end{figure*}

\section{Experiments}

\subsection{Experimental Settings}

\textbf{Datasets.} Our method is evaluated on the datasets from ActorsHQ~\cite{isik2023humanrf} and AvatarReX~\cite{zheng2023avatarrex}. There are totally $8+2$ character sequences. For setting consistency, we manually select 16 surrounding, full-body views from 160 views in ActorsHQ. 
The SMPL-X registrations for these sequences can be obtained by existing tools~\cite{EasyMocap}. 
For quantitative comparison, we make a train-test split for each sequence and the training set contains $1000$ frames. 


\textbf{Main Results.} The animation and relighting results are presented in Fig.~\ref{fig:animate}. 
We also present our reconstruction results, including reconstructed geometry, albedo color, material roughness and avatar relighting, in Fig.~\ref{fig:rec-more}. Our approach demonstrates the capability not only in synthesizing human images in novel poses but also in adapting to diverse lighting conditions. These results prove the robustness and adaptability of our method in achieving realistic and dynamic human representations with high-quality geometry and material decomposition.

\begin{figure*}[t]
  \centering
  \includegraphics[width=\textwidth]{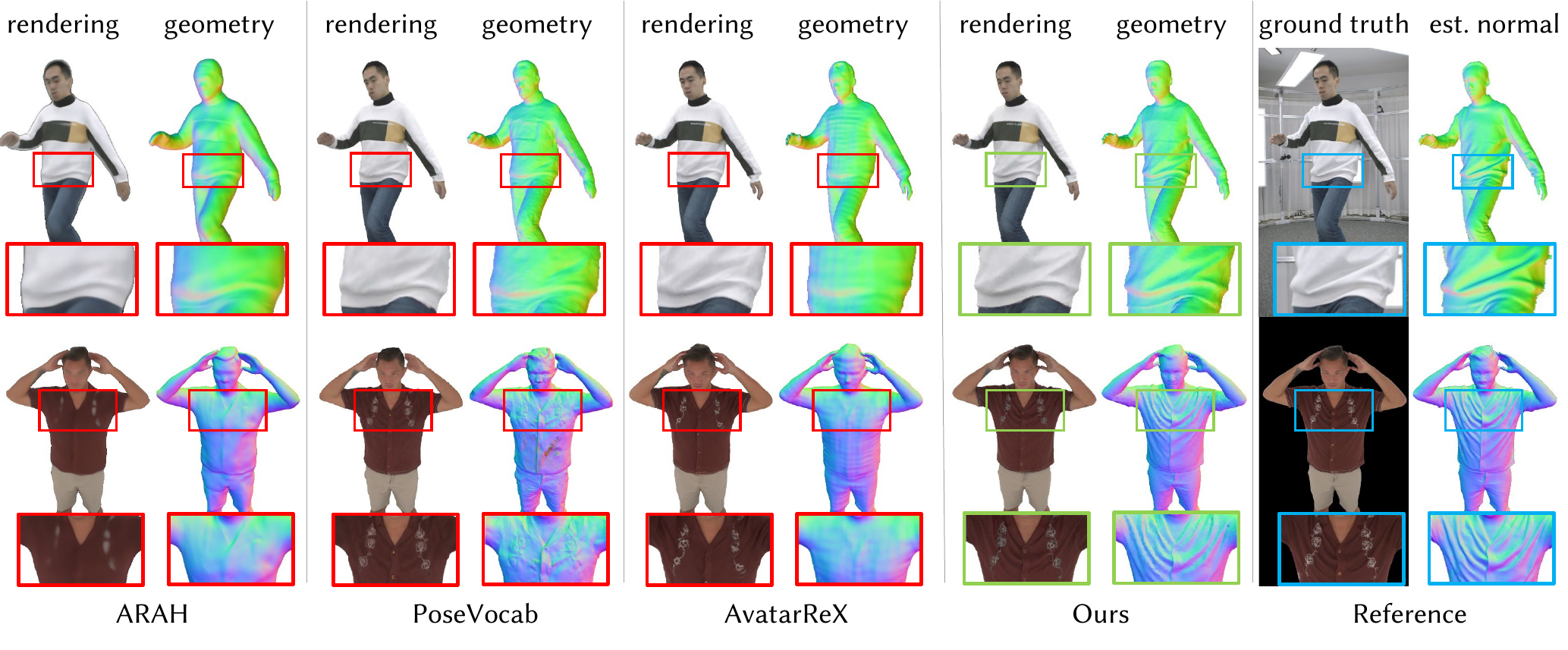}
  \caption{\textbf{Qualitative comparisons on training frame reconstructions.} Our method could reconstruct fine-grained dynamic human geometry.}
  \label{fig:train}
\end{figure*}

\begin{figure}[t]
  \centering
  \includegraphics[width=\textwidth]{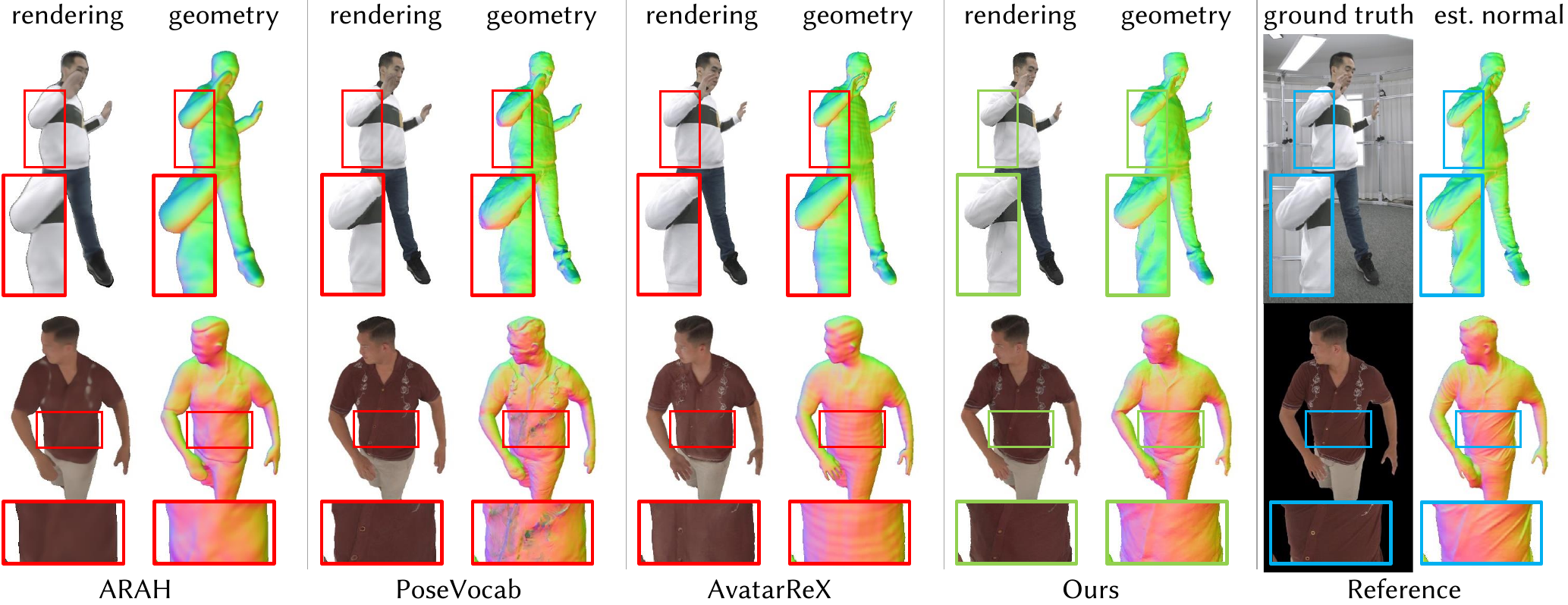}
  \caption{\textbf{Qualitative comparisons on novel pose synthesis.} Our method could not only synthesize high-quality geometry, but also more realistic appearance. }
  \label{fig:test}
\end{figure}

\subsection{Comparison}

\begin{table}[t]
\scriptsize
\centering
\caption{Quantitative comparisons on reconstructions and novel pose synthesis. The best results and the second best are highlighted in \textbf{bold} and \underline{underlined} fonts, respectively. }
\label{tab:avatar}
\begin{tabular}{@{}ccccccccc@{}}
\toprule
                & \multicolumn{4}{c}{Training Frame Reconstruction}                                           & \multicolumn{4}{c}{Novel Pose Synthesis}                                                    \\ \midrule
                & PSNR $\uparrow$       & SSIM $\uparrow$      & LPIPS $\downarrow$   & FID $\downarrow$      & PSNR $\uparrow$       & SSIM $\uparrow$      & LPIPS $\downarrow$   & FID $\downarrow$      \\ \midrule
ARAH            & $23.1071$             & $0.9390$             & $0.1232$             & $98.1130$             & $22.9917$             & $0.9368$             & $0.1254$             & $104.3341$            \\
PoseVocab       & $27.7608$             & $0.9580$             & $0.0668$             & $41.5524$             & $26.2010$             & $0.9490$             & $0.0752$             & $41.3314$             \\
AvatarReX       & $28.4015$             & $0.9607$             & $0.0579$             & $30.6214$             & $26.5853$             & $0.9497$             & $\underline{0.0684}$ & $36.1911$             \\
ours (radiance) & $\mathbf{28.9939}$    & $\mathbf{0.9646}$    & $\mathbf{0.0497}$    & $\mathbf{24.1061}$    & $\underline{26.6194}$ & $\underline{0.9498}$ & $\mathbf{0.0657}$    & $\mathbf{31.5325}$    \\
ours (PBR)      & $\underline{28.9496}$ & $\underline{0.9630}$ & $\underline{0.0559}$ & $\underline{27.0561}$ & $\mathbf{26.8526}$    & $\mathbf{0.9505}$    & $0.0693$             & $\underline{35.4671}$ \\ \bottomrule
\end{tabular}
\end{table}

\begin{figure}[t]
    \centering
    \includegraphics[width=0.8\linewidth]{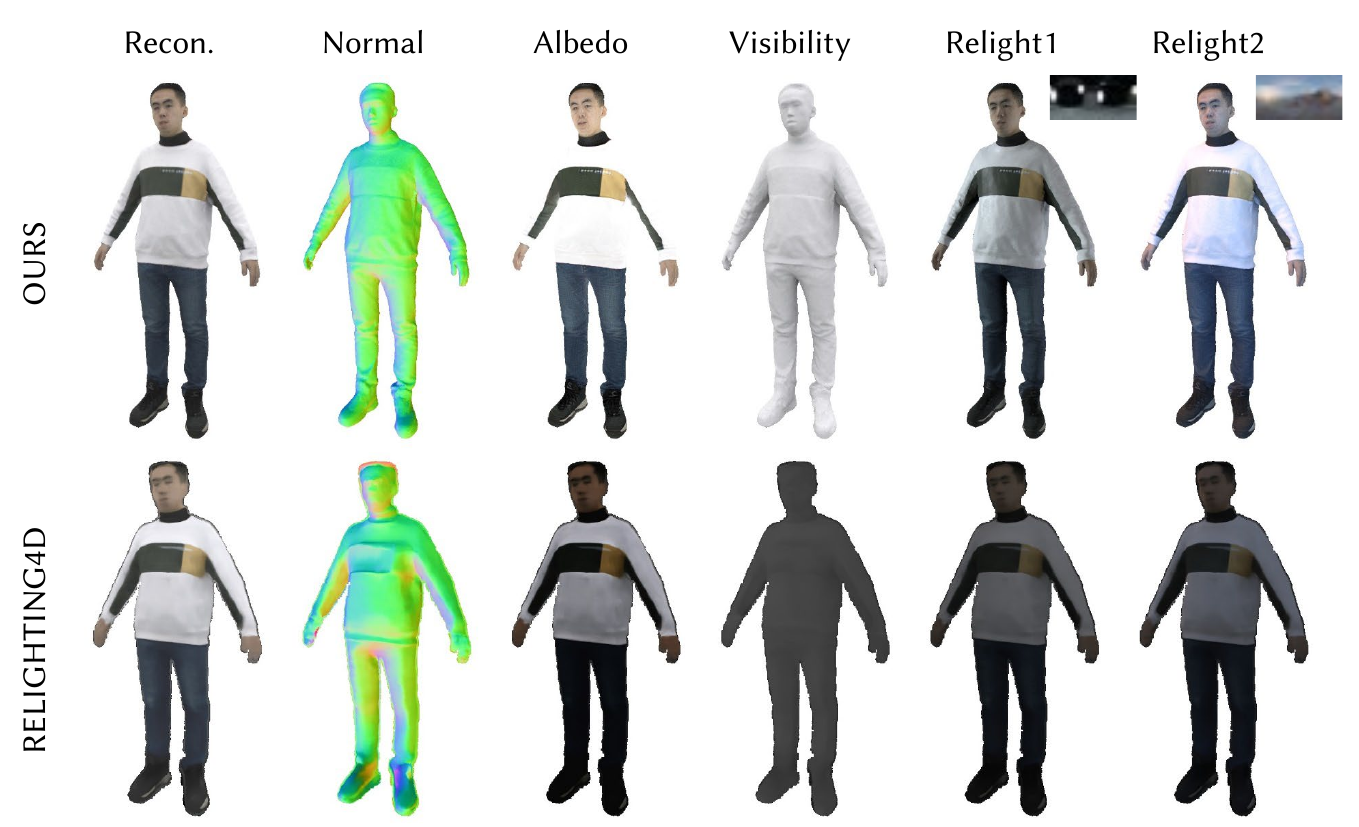}
    \caption{\textbf{Qualitative comparison} against Relighting4D~\cite{chen2022relighting4d} in terms of material decomposition and relighting fidelity.}
    \label{fig:relight}
\end{figure}

We compare our methods with recent SOTA works on neural avatars, including ARAH~\cite{wang2022arah}, PoseVocab~\cite{li2023posevocab} and AvatarReX~\cite{zheng2023avatarrex}. The results of ARAH and PoseVocab are reproduced by the public codes and the results of AvatarReX are requested from the authors. In order to have a fair and complete evaluation, we implemented a NeRF-version for our method, denoted as \textit{ours (radiance)}, alongside the primary version referred to as \textit{ours (PBR)}. Specifically, they are different only in the rendering process. In \textit{Ours(radiance)}, the surface color is directly predicted by neural networks like NeRF, instead of integrating the reflective rays on the upper hemisphere in \textit{ours (PBR)}. For quantitative comparisons on synthesized images, we use the common Peak Signal-to-Noise Ratio (PSNR), Structure Similarity Index Measure (SSIM)~\cite{wang2004image}, Learned Perceptual Image Patch Similarity (LPIPS)~\cite{zhang2018unreasonable} and Fréchet Inception Distance (FID)~\cite{heusel2017gans} as our metrics. We additionally evaluated the reconstructed geometry on the rendered normal map. The normal error with pseudo ground truth is $12.17$ (degree).

We assess the effectiveness of our approach through the evaluations on both reconstruction and novel pose synthesis. To evaluate the reconstruction quality, we present the qualitative comparisons on both the AvatarReX and ActorsHQ datasets in Fig.~\ref{fig:train}. The `geometry' column depicts the rendered normal map produced by each method, with the final column corresponding to the estimated normals derived from the raw image data. Notably, our method excels in reconstructing dynamic geometric details, such as cloth wrinkles, and capturing high-fidelity appearances. In contrast, existing approaches often rely on baking these appearance details onto a smooth human body surface.
In terms of novel pose synthesis, Fig.~\ref{fig:test} showcases the synthesized results under novel poses. Our learned human geometry demonstrates the capability of generalizing to unseen poses, and generating plausible geometric details. This capability leads to better realism in the synthesis quality achieved by our method.

 


Tab.~\ref{tab:avatar} provides the quantitative comparisons conducted on the AvatarReX dataset. Notably, the NeRF version of our method exhibits superior performance, proving that our hybrid representation surpasses all three recent approaches in human avatar modeling. Despite a minor performance decline due to rendering constraints, the PBR version still outperforms previous methods, underscoring the efficacy of our accurate geometry modeling.

To evaluate our performance on material decomposition and avatar relighting, we conduct a comparative analysis with Relighting4D~\cite{chen2022relighting4d}, one of SOTA inverse rendering pipelines designed for human motion sequences. Fig.~\ref{fig:relight} visually compares the reconstruction and relighting outcomes on one of the training frame. As shown in the figure, our method excels in learning more accurate geometry and albedo color, leading to superior results in synthesizing realistic outputs under novel lighting conditions. 
In contrast, Relighting4D struggles with both geometry reconstruction and material decomposition. 
This comparison shows the enhanced capabilities of our approach in capturing and rendering intricate details in both geometry and appearance. Due to the absence of public code, we could not compare with other previous works like \cite{iqbal2023rana, sun2023neural}. More experiments on material decomposition and relighting are presented in the supplemental material.

\begin{figure}[t]
    \centering
    \includegraphics[width=0.8\linewidth]{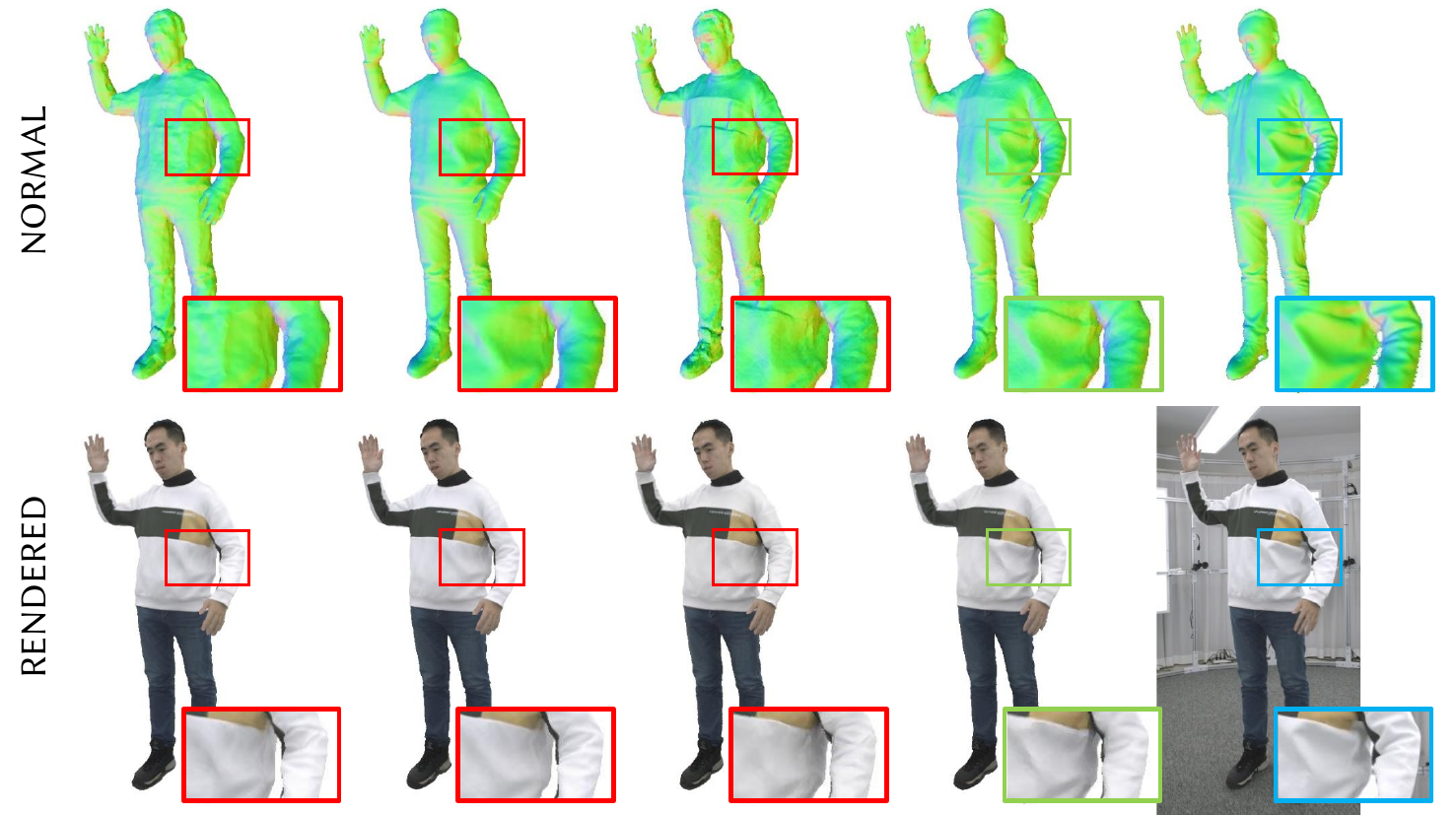}
    \caption{Ablation Studies. (a) w/o PBR and normal; (b) w/o PBR; (c) w/o normal; (d) ours; (e) reference image and estimated normal. }
    \label{fig:ablation}
\end{figure}

\subsection{Ablation Study}

In this section, we evaluate the effectiveness of PBR rendering and pseudo-normal supervision in the context of learning human geometry. Fig. \ref{fig:ablation} visually presents rendered images and normal maps for a representative training frame, illustrating the impact of each component. 
(a) Similar to other NeRF-based neural avatars, the exclusive prediction of radiance suffers from shape-radiance ambiguity. While it may achieve fair rendering quality, it often falls short in reconstructing complex geometric details.
(b) Pseudo normal ground-truth supervision encourages the generation of similar geometric structures, albeit with limitations imposed by multi-view consistency.
(c) The introduction of shadow-aware PBR proves beneficial in capturing geometric details; see the wrinkles on the arm. However, its efficacy is constrained when tasked with recovering concave structures, particularly when provided with sparse views.
(d) By leveraging the strengths of both approaches, we successfully combine the merits of the two worlds, resulting in the reconstruction of high-fidelity human geometry.

\section{Conclusion}

We introduce a novel framework for high-quality triangular human avatar modeling from multi-view videos. Our approach represents the avatar with an explicit triangular mesh, which is extracted from an implicit SDF field, complemented by a pose-dependent material field learned through the utilization of differentiable physics-based ray tracing and rendering. Moreover, we integrate 2D feature maps as well as normal supervision to further facilitate geometric detail learning. 

\begin{figure}[t]
    \centering
    \begin{subfigure}{0.67\linewidth}

        \includegraphics[width=\linewidth]{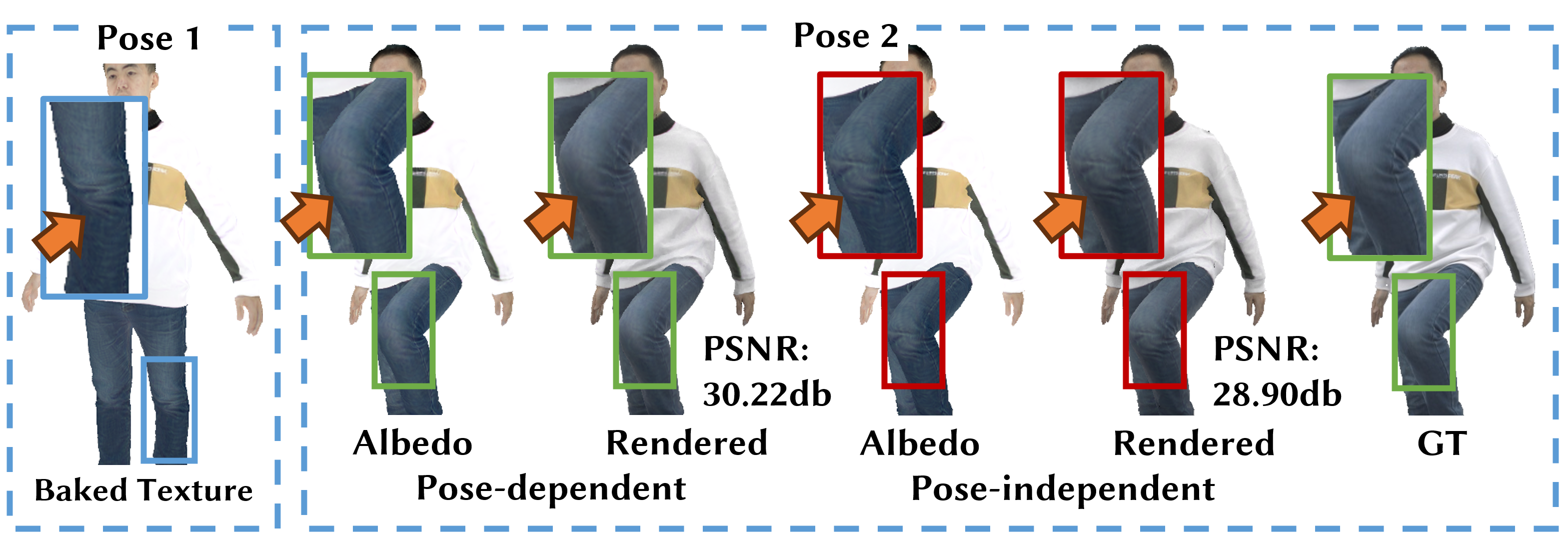}
        \caption{Pose-dependent v.s. pose-independent material fields.}
        \label{fig:pose}
    \end{subfigure}
    \hspace{5pt}
    \begin{subfigure}{0.29\linewidth}
        \includegraphics[width=\linewidth]{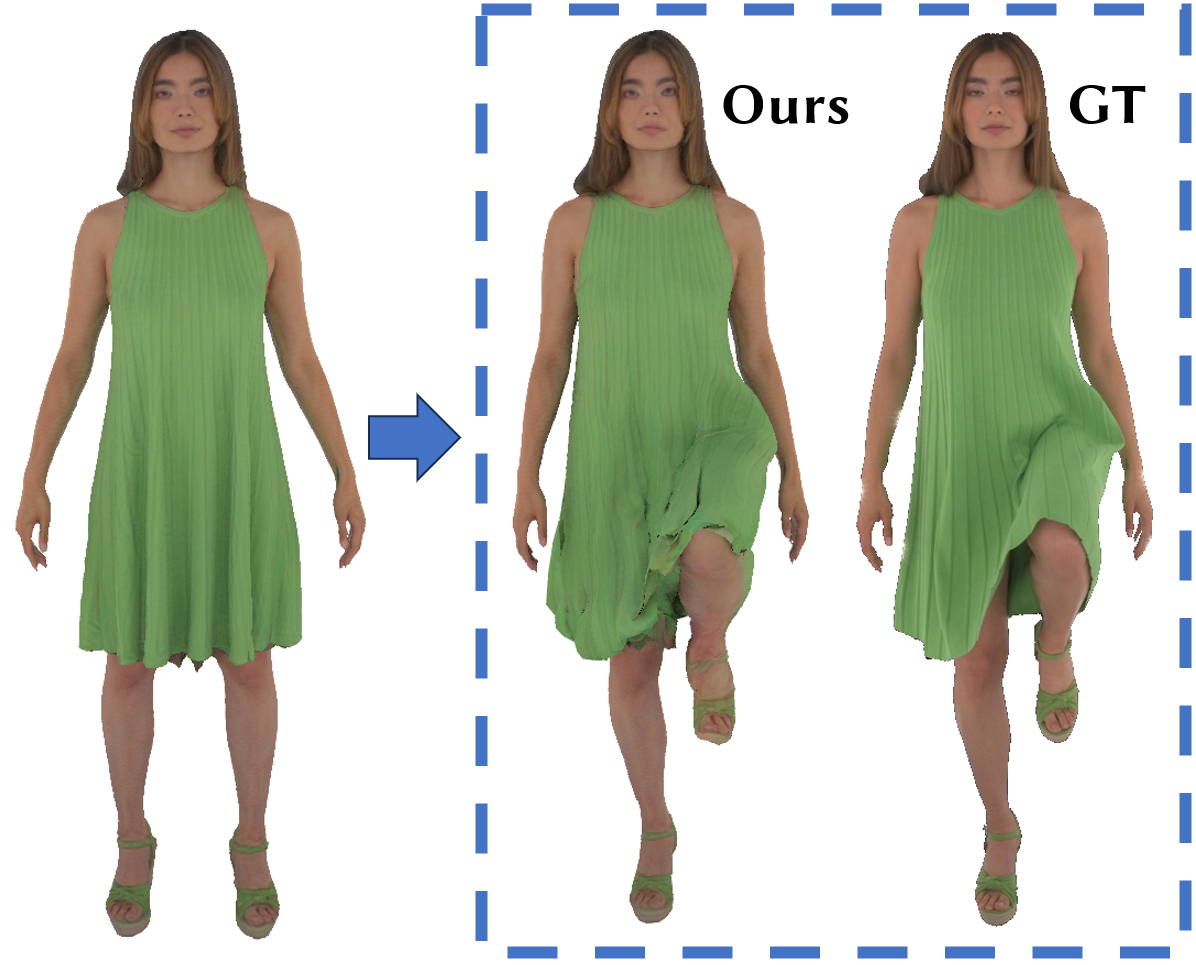}
        \caption{Failure case (long dress). }
        \label{fig:dress}
    \end{subfigure}
    \caption{Limitations of our method. (a) It's challenging to factorize all the shadows due to the limited resolution of the mesh geometry, which leads to pose-dependent shadow baking. In practice, the material field is conditioned on poses. (b) For loose garments, our method failed to learn the correct geometric deformation across different poses.}
\end{figure}

\noindent \textbf{Limitations.} 
A limitation of our work is that we use pose-dependent material to compensate the geometric errors caused by the limited resolution of the triangular mesh. This design, though practical, is not physically plausible. (Fig. \ref{fig:pose}). 
In addition, more challenging garment types like loose dresses exhibit complex non-rigid deformations, which may poses difficulties in our framework (Fig. \ref{fig:dress}). 

\noindent \textbf{Acknowledgements.} This paper is supported by National Science and Technology Major Project (2021ZD0113503) and the NSFC project No.62125107.



%
%
\bibliographystyle{splncs04}
\bibliography{reference}

\begin{thebibliography}{100}
\providecommand{\url}[1]{\texttt{#1}}
\providecommand{\urlprefix}{URL }
\providecommand{\doi}[1]{https://doi.org/#1}

\bibitem{EasyMocap}
\url{https://github.com/zju3dv/EasyMocap}

\bibitem{alldieck2018video}
Alldieck, T., Magnor, M., Xu, W., Theobalt, C., Pons-Moll, G.: Video based reconstruction of 3d people models. In: CVPR. pp. 8387--8397 (2018)

\bibitem{bagautdinov2021driving}
Bagautdinov, T., Wu, C., Simon, T., Prada, F., Shiratori, T., Wei, S.E., Xu, W., Sheikh, Y., Saragih, J.: Driving-signal aware full-body avatars. TOG  \textbf{40}(4),  1--17 (2021)

\bibitem{barron2012shape}
Barron, J.T., Malik, J.: Shape, albedo, and illumination from a single image of an unknown object. In: 2012 IEEE Conference on Computer Vision and Pattern Recognition. pp. 334--341. IEEE (2012)

\bibitem{bi2020deep}
Bi, S., Xu, Z., Sunkavalli, K., Kriegman, D., Ramamoorthi, R.: Deep 3d capture: Geometry and reflectance from sparse multi-view images. In: Proceedings of the IEEE/CVF conference on computer vision and pattern recognition. pp. 5960--5969 (2020)

\bibitem{boss2021nerd}
Boss, M., Braun, R., Jampani, V., Barron, J.T., Liu, C., Lensch, H.: Nerd: Neural reflectance decomposition from image collections. In: Proceedings of the IEEE/CVF International Conference on Computer Vision. pp. 12684--12694 (2021)

\bibitem{boss2021neural}
Boss, M., Jampani, V., Braun, R., Liu, C., Barron, J., Lensch, H.: Neural-pil: Neural pre-integrated lighting for reflectance decomposition. Advances in Neural Information Processing Systems  \textbf{34},  10691--10704 (2021)

\bibitem{disney}
Burley, B.: Physically based shading at disney (2012)

\bibitem{burov2021dynamic}
Burov, A., Nie{\ss}ner, M., Thies, J.: Dynamic surface function networks for clothed human bodies. In: ICCV. pp. 10754--10764 (2021)

\bibitem{chen2023fast}
Chen, X., Jiang, T., Song, J., Rietmann, M., Geiger, A., Black, M.J., Hilliges, O.: Fast-snarf: A fast deformer for articulated neural fields. IEEE T-PAMI  (2023)

\bibitem{chen2021snarf}
Chen, X., Zheng, Y., Black, M.J., Hilliges, O., Geiger, A.: Snarf: Differentiable forward skinning for animating non-rigid neural implicit shapes. In: ICCV. pp. 11594--11604 (2021)

\bibitem{chen2023uv}
Chen, Y., Wang, X., Chen, X., Zhang, Q., Li, X., Guo, Y., Wang, J., Wang, F.: Uv volumes for real-time rendering of editable free-view human performance. In: CVPR. pp. 16621--16631 (2023)

\bibitem{chen2022relighting4d}
Chen, Z., Liu, Z.: Relighting4d: Neural relightable human from videos. In: European Conference on Computer Vision. pp. 606--623. Springer (2022)

\bibitem{debevec2012light}
Debevec, P.: The light stages and their applications to photoreal digital actors. SIGGRAPH Asia  \textbf{2}(4), ~1--6 (2012)

\bibitem{deng2020nasa}
Deng, B., Lewis, J.P., Jeruzalski, T., Pons-Moll, G., Hinton, G., Norouzi, M., Tagliasacchi, A.: Nasa neural articulated shape approximation. In: ECCV. pp. 612--628. Springer (2020)

\bibitem{dong2022totalselfscan}
Dong, J., Fang, Q., Guo, Y., Peng, S., Shuai, Q., Zhou, X., Bao, H.: Totalselfscan: Learning full-body avatars from self-portrait videos of faces, hands, and bodies. In: NeurIPS (2022)

\bibitem{dong2014appearance}
Dong, Y., Chen, G., Peers, P., Zhang, J., Tong, X.: Appearance-from-motion: Recovering spatially varying surface reflectance under unknown lighting. ACM Transactions on Graphics (TOG)  \textbf{33}(6),  1--12 (2014)

\bibitem{dong2022pina}
Dong, Z., Guo, C., Song, J., Chen, X., Geiger, A., Hilliges, O.: Pina: Learning a personalized implicit neural avatar from a single rgb-d video sequence. In: CVPR (2022)

\bibitem{Feng2022scarf}
Feng, Y., Yang, J., Pollefeys, M., Black, M.J., Bolkart, T.: Capturing and animation of body and clothing from monocular video. In: SIGGRAPH Asia 2022 Conference Proceedings. SA '22 (2022)

\bibitem{goldman2009shape}
Goldman, D.B., Curless, B., Hertzmann, A., Seitz, S.M.: Shape and spatially-varying brdfs from photometric stereo. IEEE Transactions on Pattern Analysis and Machine Intelligence  \textbf{32}(6),  1060--1071 (2009)

\bibitem{gropp2020implicit}
Gropp, A., Yariv, L., Haim, N., Atzmon, M., Lipman, Y.: Implicit geometric regularization for learning shapes. In: ICML. pp. 3789--3799. PMLR (2020)

\bibitem{guan2012drape}
Guan, P., Reiss, L., Hirshberg, D.A., Weiss, A., Black, M.J.: Drape: Dressing any person. TOG  \textbf{31}(4),  1--10 (2012)

\bibitem{guo2023vid2avatar}
Guo, C., Jiang, T., Chen, X., Song, J., Hilliges, O.: Vid2avatar: 3d avatar reconstruction from videos in the wild via self-supervised scene decomposition. In: CVPR. pp. 12858--12868 (2023)

\bibitem{guo2019relightables}
Guo, K., Lincoln, P., Davidson, P., Busch, J., Yu, X., Whalen, M., Harvey, G., Orts-Escolano, S., Pandey, R., Dourgarian, J., et~al.: The relightables: Volumetric performance capture of humans with realistic relighting. ACM Transactions on Graphics (ToG)  \textbf{38}(6),  1--19 (2019)

\bibitem{habermann2023hdhumans}
Habermann, M., Liu, L., Xu, W., Pons-Moll, G., Zollhoefer, M., Theobalt, C.: Hdhumans: A hybrid approach for high-fidelity digital humans. ACM SCA  \textbf{6}(3),  1--23 (2023)

\bibitem{habermann2021real}
Habermann, M., Liu, L., Xu, W., Zollhoefer, M., Pons-Moll, G., Theobalt, C.: Real-time deep dynamic characters. TOG  \textbf{40}(4),  1--16 (2021)

\bibitem{hasselgren2022shape}
Hasselgren, J., Hofmann, N., Munkberg, J.: Shape, light, and material decomposition from images using monte carlo rendering and denoising. Advances in Neural Information Processing Systems  \textbf{35},  22856--22869 (2022)

\bibitem{heusel2017gans}
Heusel, M., Ramsauer, H., Unterthiner, T., Nessler, B., Hochreiter, S.: Gans trained by a two time-scale update rule converge to a local nash equilibrium. NeurIPS  \textbf{30} (2017)

\bibitem{ho2023learning}
Ho, H.I., Xue, L., Song, J., Hilliges, O.: Learning locally editable virtual humans. In: CVPR. pp. 21024--21035 (2023)

\bibitem{hong2021stereopifu}
Hong, Y., Zhang, J., Jiang, B., Guo, Y., Liu, L., Bao, H.: Stereopifu: Depth aware clothed human digitization via stereo vision. In: CVPR. pp. 535--545 (2021)

\bibitem{hu2023gauhuman}
Hu, S., Liu, Z.: Gauhuman: Articulated gaussian splatting from monocular human videos. arXiv preprint arXiv:  (2023)

\bibitem{isik2023humanrf}
I\c{s}{\i}k, M., Rünz, M., Georgopoulos, M., Khakhulin, T., Starck, J., Agapito, L., Nießner, M.: Humanrf: High-fidelity neural radiance fields for humans in motion. TOG  \textbf{42}(4),  1--12 (2023)

\bibitem{iqbal2023rana}
Iqbal, U., Caliskan, A., Nagano, K., Khamis, S., Molchanov, P., Kautz, J.: Rana: Relightable articulated neural avatars. In: Proceedings of the IEEE/CVF International Conference on Computer Vision. pp. 23142--23153 (2023)

\bibitem{Iwase_2023}
Iwase, S., Saito, S., Simon, T., Lombardi, S., Timur, B., Joshi, R., Prada, F., Shiratori, T., Sheikh, Y., Saragih, J.: Relightablehands: Efficient neural relighting of articulated hand models. In: CVPR (2023)

\bibitem{ji2022relight}
Ji, C., Yu, T., Guo, K., Liu, J., Liu, Y.: Geometry-aware single-image full-body human relighting  (October 2022)

\bibitem{Jiang2023nerffacelighting}
Jiang, K., Chen, S.Y., Fu, H., Gao, L.: Nerffacelighting: Implicit and disentangled face lighting representation leveraging generative prior in neural radiance fields. ACM Transactions on Graphics (TOG)  (2023)

\bibitem{jiang2023instantavatar}
Jiang, T., Chen, X., Song, J., Hilliges, O.: Instantavatar: Learning avatars from monocular video in 60 seconds. In: CVPR. pp. 16922--16932 (2023)

\bibitem{jiang2022neuman}
Jiang, W., Yi, K.M., Samei, G., Tuzel, O., Ranjan, A.: Neuman: Neural human radiance field from a single video. In: ECCV. pp. 402--418. Springer (2022)

\bibitem{jin2023tensoir}
Jin, H., Liu, I., Xu, P., Zhang, X., Han, S., Bi, S., Zhou, X., Xu, Z., Su, H.: Tensoir: Tensorial inverse rendering. In: Proceedings of the IEEE/CVF Conference on Computer Vision and Pattern Recognition. pp. 165--174 (2023)

\bibitem{kajiya1986rendering}
Kajiya, J.T.: The rendering equation. In: Proceedings of the 13th annual conference on Computer graphics and interactive techniques. pp. 143--150 (1986)

\bibitem{Kanamori2018relightinghumans}
Kanamori, Y., Endo, Y.: Relighting humans: occlusion-aware inverse rendering for full-body human images. ACM Trans. Graph.  \textbf{37}(6) (dec 2018)

\bibitem{kerbl2023gaussian}
Kerbl, B., Kopanas, G., Leimk{\"u}hler, T., Drettakis, G.: 3d gaussian splatting for real-time radiance field rendering. TOG  \textbf{42}(4),  1--14 (2023)

\bibitem{kim2022laplacianfusion}
Kim, H., Nam, H., Kim, J., Park, J., Lee, S.: Laplacianfusion: Detailed 3d clothed-human body reconstruction. ACM Transactions on Graphics (TOG)  \textbf{41}(6),  1--14 (2022)

\bibitem{adam}
Kingma, D.P., Ba, J.: Adam: A method for stochastic optimization. In: ICLR (2015)

\bibitem{kocabas2023hugs}
Kocabas, M., Chang, J.H.R., Gabriel, J., Tuzel, O., Ranjan, A.: Hugs: Human gaussian splats (2023)

\bibitem{Kwon2023deliffas}
Kwon, Y., Liu, L., Fuchs, H., Habermann, M., Theobalt, C.: Deliffas: Deformable light fields for fast avatar synthesis. In: NeurIPS (2023)

\bibitem{Lagunas2021humanrelighting}
Lagunas, M., Sun, X., Yang, J., Villegas, R., Zhang, J., Shu, Z., Masia, B., Gutierrez, D.: Single-image full-body human relighting. In: Eurographics Symposium on Rendering (EGSR). The Eurographics Association (2021). \doi{10.2312/sr.20211301}

\bibitem{lawrence2004efficient}
Lawrence, J., Rusinkiewicz, S., Ramamoorthi, R.: Efficient brdf importance sampling using a factored representation. ACM Transactions on Graphics (ToG)  \textbf{23}(3),  496--505 (2004)

\bibitem{legendre2019deeplight}
LeGendre, C., Ma, W.C., Fyffe, G., Flynn, J., Charbonnel, L., Busch, J., Debevec, P.: Deeplight: Learning illumination for unconstrained mobile mixed reality. In: Proceedings of the IEEE/CVF Conference on Computer Vision and Pattern Recognition. pp. 5918--5928 (2019)

\bibitem{li2022tava}
Li, R., Tanke, J., Vo, M., Zollh{\"o}fer, M., Gall, J., Kanazawa, A., Lassner, C.: Tava: Template-free animatable volumetric actors. In: ECCV. pp. 419--436. Springer (2022)

\bibitem{li2023posevocab}
Li, Z., Zheng, Z., Liu, Y., Zhou, B., Liu, Y.: Posevocab: Learning joint-structured pose embeddings for human avatar modeling. In: ACM SIGGRAPH Conference Proceedings (2023)

\bibitem{li2023animatable}
Li, Z., Zheng, Z., Wang, L., Liu, Y.: Animatable gaussians: Learning pose-dependent gaussian maps for high-fidelity human avatar modeling. arXiv preprint arXiv:2311.16096  (2023)

\bibitem{li2022avatarcap}
Li, Z., Zheng, Z., Zhang, H., Ji, C., Liu, Y.: Avatarcap: Animatable avatar conditioned monocular human volumetric capture. In: ECCV. pp. 322--341. Springer (2022)

\bibitem{li2018learning}
Li, Z., Xu, Z., Ramamoorthi, R., Sunkavalli, K., Chandraker, M.: Learning to reconstruct shape and spatially-varying reflectance from a single image. ACM Transactions on Graphics (TOG)  \textbf{37}(6),  1--11 (2018)

\bibitem{lin2024layga}
Lin, S., Li, Z., Su, Z., Zheng, Z., Zhang, H., Liu, Y.: Layga: Layered gaussian avatars for animatable clothing transfer. In: SIGGRAPH Conference Papers (2024)

\bibitem{lin2022learning}
Lin, S., Zhang, H., Zheng, Z., Shao, R., Liu, Y.: Learning implicit templates for point-based clothed human modeling. In: ECCV. pp. 210--228. Springer (2022)

\bibitem{lin2024relightable}
Lin, W., Zheng, C., Yong, J.H., Xu, F.: Relightable and animatable neural avatars from videos. AAAI  (2024)

\bibitem{lipson2021raftstereo}
Lipson, L., Teed, Z., Deng, J.: Raft-stereo: Multilevel recurrent field transforms for stereo matching. In: International Conference on 3D Vision (3DV) (2021)

\bibitem{liu2021neural}
Liu, L., Habermann, M., Rudnev, V., Sarkar, K., Gu, J., Theobalt, C.: Neural actor: Neural free-view synthesis of human actors with pose control. TOG  \textbf{40}(6),  1--16 (2021)

\bibitem{loper2015smpl}
Loper, M., Mahmood, N., Romero, J., Pons-Moll, G., Black, M.J.: Smpl: A skinned multi-person linear model. TOG  \textbf{34}(6),  1--16 (2015)

\bibitem{ma2021scale}
Ma, Q., Saito, S., Yang, J., Tang, S., Black, M.J.: Scale: Modeling clothed humans with a surface codec of articulated local elements. In: CVPR. pp. 16082--16093 (2021)

\bibitem{ma2021power}
Ma, Q., Yang, J., Tang, S., Black, M.J.: The power of points for modeling humans in clothing. In: ICCV. pp. 10974--10984 (2021)

\bibitem{mescheder2019occupancy}
Mescheder, L., Oechsle, M., Niemeyer, M., Nowozin, S., Geiger, A.: Occupancy networks: Learning 3d reconstruction in function space. In: CVPR. pp. 4460--4470 (2019)

\bibitem{mihajlovic2021leap}
Mihajlovic, M., Zhang, Y., Black, M.J., Tang, S.: Leap: Learning articulated occupancy of people. In: CVPR. pp. 10461--10471 (2021)

\bibitem{mildenhall2020nerf}
Mildenhall, B., Srinivasan, P.P., Tancik, M., Barron, J.T., Ramamoorthi, R., Ng, R.: Nerf: Representing scenes as neural radiance fields for view synthesis. In: ECCV. pp. 405--421. Springer (2020)

\bibitem{Munkberg2022nvdiffrec}
Munkberg, J., Hasselgren, J., Shen, T., Gao, J., Chen, W., Evans, A., M\"uller, T., Fidler, S.: {Extracting Triangular 3D Models, Materials, and Lighting From Images}. In: Proceedings of the IEEE/CVF Conference on Computer Vision and Pattern Recognition (CVPR). pp. 8280--8290 (June 2022)

\bibitem{nam2018practical}
Nam, G., Lee, J.H., Gutierrez, D., Kim, M.H.: Practical svbrdf acquisition of 3d objects with unstructured flash photography. ACM Transactions on Graphics (TOG)  \textbf{37}(6),  1--12 (2018)

\bibitem{zhu2023ash}
Pang, H., Zhu, H., Kortylewski, A., Theobalt, C., Habermann, M.: Ash: Animatable gaussian splats for efficient and photoreal human rendering  (2023)

\bibitem{park2019deepsdf}
Park, J.J., Florence, P., Straub, J., Newcombe, R., Lovegrove, S.: Deepsdf: Learning continuous signed distance functions for shape representation. In: CVPR. pp. 165--174 (2019)

\bibitem{peng2022selfnerf}
Peng, B., Hu, J., Zhou, J., Zhang, J.: Selfnerf: Fast training nerf for human from monocular self-rotating video. arXiv preprint arXiv:2210.01651  (2022)

\bibitem{peng2021animatable}
Peng, S., Dong, J., Wang, Q., Zhang, S., Shuai, Q., Zhou, X., Bao, H.: Animatable neural radiance fields for modeling dynamic human bodies. In: ICCV. pp. 14314--14323 (2021)

\bibitem{peng2022animatable}
Peng, S., Zhang, S., Xu, Z., Geng, C., Jiang, B., Bao, H., Zhou, X.: Animatable neural implicit surfaces for creating avatars from videos. arXiv preprint arXiv:2203.08133  (2022)

\bibitem{remelli2022drivable}
Remelli, E., Bagautdinov, T., Saito, S., Wu, C., Simon, T., Wei, S.E., Guo, K., Cao, Z., Prada, F., Saragih, J., et~al.: Drivable volumetric avatars using texel-aligned features. In: ACM SIGGRAPH 2022 Conference Proceedings. pp.~1--9 (2022)

\bibitem{saito2020pifuhd}
Saito, S., Simon, T., Saragih, J., Joo, H.: Pifuhd: Multi-level pixel-aligned implicit function for high-resolution 3d human digitization. In: CVPR (June 2020)

\bibitem{saito2021scanimate}
Saito, S., Yang, J., Ma, Q., Black, M.J.: Scanimate: Weakly supervised learning of skinned clothed avatar networks. In: CVPR. pp. 2886--2897 (2021)

\bibitem{shao2021doublefield}
Shao, R., Zhang, H., Zhang, H., Chen, M., Cao, Y., Yu, T., Liu, Y.: Doublefield: Bridging the neural surface and radiance fields for high-fidelity human reconstruction and rendering. In: CVPR (2022)

\bibitem{shen2023xavatar}
Shen, K., Guo, C., Kaufmann, M., Zarate, J.J., Valentin, J., Song, J., Hilliges, O.: X-avatar: Expressive human avatars. In: CVPR. pp. 16911--16921 (2023)

\bibitem{shen2021dmtet}
Shen, T., Gao, J., Yin, K., Liu, M.Y., Fidler, S.: Deep marching tetrahedra: a hybrid representation for high-resolution 3d shape synthesis. In: Advances in Neural Information Processing Systems (NeurIPS) (2021)

\bibitem{srinivasan2021nerv}
Srinivasan, P.P., Deng, B., Zhang, X., Tancik, M., Mildenhall, B., Barron, J.T.: Nerv: Neural reflectance and visibility fields for relighting and view synthesis. In: Proceedings of the IEEE/CVF Conference on Computer Vision and Pattern Recognition. pp. 7495--7504 (2021)

\bibitem{stoll2010video}
Stoll, C., Gall, J., De~Aguiar, E., Thrun, S., Theobalt, C.: Video-based reconstruction of animatable human characters. TOG  \textbf{29}(6),  1--10 (2010)

\bibitem{su2023npc}
Su, S.Y., Bagautdinov, T., Rhodin, H.: Npc: Neural point characters from video. In: ICCV (2023)

\bibitem{su2021a-nerf}
Su, S.Y., Yu, F., Zollh{\"o}fer, M., Rhodin, H.: A-nerf: Articulated neural radiance fields for learning human shape, appearance, and pose. NeurIPS  \textbf{34},  12278--12291 (2021)

\bibitem{sun2023neural}
Sun, W., Che, Y., Huang, H., Guo, Y.: Neural reconstruction of relightable human model from monocular video. In: Proceedings of the IEEE/CVF International Conference on Computer Vision. pp. 397--407 (2023)

\bibitem{tajima2021RHW}
Tajima, D., Kanamori, Y., Endo, Y.: Relighting humans in the wild: Monocular full-body human relighting with domain adaptation. Computer Graphics Forum (Proc. of Pacific Graphics 2021)  \textbf{40}(7),  205--216 (2021)

\bibitem{tancik2020fourier}
Tancik, M., Srinivasan, P., Mildenhall, B., Fridovich-Keil, S., Raghavan, N., Singhal, U., Ramamoorthi, R., Barron, J., Ng, R.: Fourier features let networks learn high frequency functions in low dimensional domains. NeurIPS  \textbf{33},  7537--7547 (2020)

\bibitem{te2022neural}
Te, G., Li, X., Li, X., Wang, J., Hu, W., Lu, Y.: Neural capture of animatable 3d human from monocular video. In: ECCV. pp. 275--291. Springer (2022)

\bibitem{tiwari2021neural}
Tiwari, G., Sarafianos, N., Tung, T., Pons-Moll, G.: Neural-gif: Neural generalized implicit functions for animating people in clothing. In: ICCV. pp. 11708--11718 (2021)

\bibitem{WangCVPR2024IntrinsicAvatar}
Wang, S., Anti\'{c}, B., Geiger, A., Tang, S.: Intrinsicavatar: Physically based inverse rendering of dynamic humans from monocular videos via explicit ray tracing. In: Proceedings IEEE Conf. on Computer Vision and Pattern Recognition (CVPR) (2024)

\bibitem{wang2021metaavatar}
Wang, S., Mihajlovic, M., Ma, Q., Geiger, A., Tang, S.: Metaavatar: Learning animatable clothed human models from few depth images. NeurIPS  \textbf{34} (2021)

\bibitem{wang2022arah}
Wang, S., Schwarz, K., Geiger, A., Tang, S.: Arah: Animatable volume rendering of articulated human sdfs. In: ECCV. pp. 1--19. Springer (2022)

\bibitem{wang2023sunstage}
Wang, Y., Holynski, A., Zhang, X., Zhang, X.: Sunstage: Portrait reconstruction and relighting using the sun as a light stage. In: Proceedings of the IEEE/CVF Conference on Computer Vision and Pattern Recognition. pp. 20792--20802 (2023)

\bibitem{wang2004image}
Wang, Z., Bovik, A.C., Sheikh, H.R., Simoncelli, E.P.: Image quality assessment: from error visibility to structural similarity. IEEE T-IP  \textbf{13}(4),  600--612 (2004)

\bibitem{weng2022humannerf}
Weng, C.Y., Curless, B., Srinivasan, P.P., Barron, J.T., Kemelmacher-Shlizerman, I.: Humannerf: Free-viewpoint rendering of moving people from monocular video. In: CVPR. pp. 16210--16220 (2022)

\bibitem{xiang2022dressing}
Xiang, D., Bagautdinov, T., Stuyck, T., Prada, F., Romero, J., Xu, W., Saito, S., Guo, J., Smith, B., Shiratori, T., et~al.: Dressing avatars: Deep photorealistic appearance for physically simulated clothing. TOG  \textbf{41}(6),  1--15 (2022)

\bibitem{xiang2021modeling}
Xiang, D., Prada, F., Bagautdinov, T., Xu, W., Dong, Y., Wen, H., Hodgins, J., Wu, C.: Modeling clothing as a separate layer for an animatable human avatar. TOG  \textbf{40}(6),  1--15 (2021)

\bibitem{xu2011video}
Xu, F., Liu, Y., Stoll, C., Tompkin, J., Bharaj, G., Dai, Q., Seidel, H.P., Kautz, J., Theobalt, C.: Video-based characters: creating new human performances from a multi-view video database. TOG  \textbf{30}(4),  1--10 (2011)

\bibitem{xu2023relightable}
Xu, Z., Peng, S., Geng, C., Mou, L., Yan, Z., Sun, J., Bao, H., Zhou, X.: Relightable and animatable neural avatar from sparse-view video. In: Proceedings IEEE Conf. on Computer Vision and Pattern Recognition (CVPR) (2024)

\bibitem{yariv2021volume}
Yariv, L., Gu, J., Kasten, Y., Lipman, Y.: Volume rendering of neural implicit surfaces. NeurIPS  \textbf{34},  4805--4815 (2021)

\bibitem{yu2021function4d}
Yu, T., Zheng, Z., Guo, K., Liu, P., Dai, Q., Liu, Y.: Function4d: Real-time human volumetric capture from very sparse consumer rgbd sensors. In: CVPR. pp. 5746--5756 (2021)

\bibitem{zhang2023closet}
Zhang, H., Lin, S., Shao, R., Zhang, Y., Zheng, Z., Huang, H., Guo, Y., Liu, Y.: Closet: Modeling clothed humans on continuous surface with explicit template decomposition. In: CVPR (2023)

\bibitem{zhang2021physg}
Zhang, K., Luan, F., Wang, Q., Bala, K., Snavely, N.: Physg: Inverse rendering with spherical gaussians for physics-based material editing and relighting. In: Proceedings of the IEEE/CVF Conference on Computer Vision and Pattern Recognition. pp. 5453--5462 (2021)

\bibitem{zhang2018unreasonable}
Zhang, R., Isola, P., Efros, A.A., Shechtman, E., Wang, O.: The unreasonable effectiveness of deep features as a perceptual metric. In: CVPR. pp. 586--595 (2018)

\bibitem{zhang2021nerfactor}
Zhang, X., Srinivasan, P.P., Deng, B., Debevec, P., Freeman, W.T., Barron, J.T.: Nerfactor: Neural factorization of shape and reflectance under an unknown illumination. ACM Transactions on Graphics (ToG)  \textbf{40}(6),  1--18 (2021)

\bibitem{zhao2022high}
Zhao, H., Zhang, J., Lai, Y.K., Zheng, Z., Xie, Y., Liu, Y., Li, K.: High-fidelity human avatars from a single rgb camera. In: CVPR. pp. 15904--15913 (2022)

\bibitem{zheng2023pointavatar}
Zheng, Y., Yifan, W., Wetzstein, G., Black, M.J., Hilliges, O.: Pointavatar: Deformable point-based head avatars from videos. In: CVPR. pp. 21057--21067 (2023)

\bibitem{zheng2022structured}
Zheng, Z., Huang, H., Yu, T., Zhang, H., Guo, Y., Liu, Y.: Structured local radiance fields for human avatar modeling. In: CVPR. pp. 15893--15903 (2022)

\bibitem{zheng2023avatarrex}
Zheng, Z., Zhao, X., Zhang, H., Liu, B., Liu, Y.: Avatarrex: Real-time expressive full-body avatars. TOG  \textbf{42}(4) (2023). \doi{10.1145/3592101}

\bibitem{Zielonka2023Drivable3D}
Zielonka, W., Bagautdinov, T., Saito, S., Zollhöfer, M., Thies, J., Romero, J.: Drivable 3d gaussian avatars  (2023)

\end{thebibliography}

\clearpage

\appendix
\setcounter{table}{0}
\setcounter{figure}{0}
\renewcommand{\thetable}{\Alph{table}}
\renewcommand{\thefigure}{\Alph{figure}}

\section{Implementation Details}
\label{sec:impl_details}

Our learning objective 
\begin{equation}
    \mathcal L = \mathcal L_\mathrm{img} + \mathcal L_\mathrm{reg} \label{eq:loss}
\end{equation}
consists of the photometric loss between rendered image $\hat {\mathbf I}$/normal map $\hat {\mathbf N}$ and the input image $\mathbf I$/estimated normal $\mathbf N$, plus the regularization terms for our implicit SDFs, pose-dependent vertex offsets, materials and lighting.
\begin{equation}
    \mathcal L_\mathrm{img} = \left\|\hat {\mathbf{I}} - \gamma_\mathrm{tone}^{-1}(\mathbf I)\right\|_1 + \left\|\hat {\mathbf{N}} - \mathbf N\right\|_1 
    +\lambda_\mathrm{LPIPS}\mathcal L_\mathrm{LPIPS} \left(\gamma_\mathrm{tone}(\hat{\mathbf I}), \mathbf I\right),
\end{equation}
\begin{equation}
    \mathcal L_\mathrm{reg} = \lambda_\mathrm{SDF}\mathcal L_\mathrm{SDF} + \lambda_\mathrm{offset}\mathcal L_\mathrm{offset} +\lambda_\mathrm{mat}\mathcal L_\mathrm{mat} + \lambda_\mathrm{light}\mathcal L_\mathrm{light},
\end{equation}
where $\mathcal L_\mathrm{LPIPS}$ is the perceptual loss~\cite{zhang2018unreasonable}, $\gamma_\mathrm{tone}$ is the tone mapping function to map the rendered image from linear color space to sRGB color space, and $\lambda_\mathrm{LPIPS}, \lambda_\mathrm{SDF}, \lambda_\mathrm{mat}, \lambda_\mathrm{offset}, \lambda_\mathrm{light}$ are balancing coefficients,
\begin{equation}
    \mathcal L_\mathrm{SDF} = \sum_{x\in\mathbb R^3} \|\nabla_x \mathcal S(x) - 1\|^2,\ \mathcal L_\mathrm{offset} = \frac{1}{N_c}\sum_{n=1}^{N_c} \|\Delta \mathbf{v}_n\|^2
\end{equation}
encourages the base mesh to be smooth~\cite{gropp2020implicit} and fit the clothed human as much as possible, and
\begin{equation}
    \mathcal L_\mathrm{mat} = \sum_{\mathbf v \in \mathcal M, k\in \{k_d, k_s\}} |k(\mathbf v) - k(\mathbf v + \epsilon)|,
\end{equation}
\begin{equation}
    \mathcal L_\mathrm{light} = \frac{1}{3}\sum_{c\in \{r,g,b\}}\left|L_{\mathrm{env},c} - \frac{1}{3}\sum_{c\in \{r,g,b\}} L_\mathrm{env,c}\right|
\end{equation}
are used to regularize spatially-smooth material and low-frequency lighting~\cite{Munkberg2022nvdiffrec}.

    Besides dataset preprocessing like SMPL-X registration and stereo-based normal estimation, our avatar modeling pipeline is completely end-to-end, with the supervision signal from Equation \ref{eq:loss}. To ensure that the geometric details can be generated, the resolution of our tetrahedral grid is set 256 (Every edge in the grid has length $1/128$ m). The tetrahedral grid is mainly defined around the SMPL-X body shape to avoid SDF queries in unnecessary regions. The extracted human mesh has $\sim35$k vertices and $\sim70$k triangles. The SDF field is implemented and initialized the same as in VolSDF~\cite{yariv2021volume}. During training, the balancing loss coefficients are set $\lambda_\mathrm{LPIPS}=0.1$, $\lambda_\mathrm{SDF}=0.01$, $\lambda_\mathrm{mat}=0.3$ for diffuse albedo and $0.05$ for surface roughness. $\lambda_\mathrm{offset}$ decreases linearly from $1000$ to $10$ to learn a meaningful base mesh at the early steps.

\begin{table}[t]
\caption{Comparisons on training/inference time with other SOTA methods on neural avatars.}
\label{tab:time}
\scriptsize
\centering
\begin{tabular}{@{}cccccccc@{}}
\toprule
                                                                            & Ours      & \begin{tabular}[c]{@{}c@{}}PoseVocab\\ \cite{li2023posevocab}\end{tabular}      & \begin{tabular}[c]{@{}c@{}}AvatarReX\\ \cite{zheng2023avatarrex}\end{tabular}& \begin{tabular}[c]{@{}c@{}}Animatable\\ Gaussians\cite{li2023animatable}\end{tabular}                                         & \begin{tabular}[c]{@{}c@{}}Xu \etal\\ \cite{xu2023relightable}\end{tabular}& \begin{tabular}[c]{@{}c@{}}Lin \etal\\ \cite{lin2024relightable}\end{tabular}     & \begin{tabular}[c]{@{}c@{}}Wang \etal\\ \cite{WangCVPR2024IntrinsicAvatar}\end{tabular}                                                     \\ \midrule
Relightable?                                                                & $\checkmark$       &                &           &                                                            & $\checkmark$ & $\checkmark$     & $\checkmark$                                                      \\ \midrule
\begin{tabular}[c]{@{}c@{}}Training Time\\ ($\sim$100 frames)\end{tabular}  & $\sim$3h  &                &           &                                                            &     & 2.5 days & \begin{tabular}[c]{@{}c@{}}4h\\ (monocular)\end{tabular} \\
\begin{tabular}[c]{@{}c@{}}Training Time\\ ($\sim$1000 frames)\end{tabular} & $\sim$16h & 1.5$\sim$2 days & 2 days     & \begin{tabular}[c]{@{}c@{}}2 days\\ (RTX 4090)\end{tabular} & 30h &         &                                                          \\ \midrule
\begin{tabular}[c]{@{}c@{}}Inference Time\\ (per image)\end{tabular}        & 180ms     & 3s             & 30s       & \begin{tabular}[c]{@{}c@{}}100ms\\ (1024$\times$1024)\end{tabular}                                                      & 5s  & 40s     & 20s                                                      \\ \bottomrule
\end{tabular}
\end{table}

    Our model is trained on single NVIDIA RTX 3090 GPU using Adam~\cite{adam} optimizer. It takes 100k steps and around 16 hours to converge.
    At the inference time, we sample $64$ reflective rays without denoiser for more realistic and accurate rendering. It takes only 180ms (35ms to generate posed mesh + 145ms to render) to render an image of $512\times 512$ resolution, in contrast to 40s in \cite{lin2024relightable}, $\sim$20s in \cite{WangCVPR2024IntrinsicAvatar}, 5s in \cite{xu2023relightable} and 50s in \cite{chen2022relighting4d}, proving the effectiveness of our mesh-based representation. More comparisons with other SOTA neural avatars are shown in Table \ref{tab:time}. Existing works using 3DGS~\cite{li2023animatable} could not produce accurate dynamic geometries nor support relighting under novel environments. Note that the former 35ms is inevitable to produce pose-dependent dynamic details using neural networks (even if using 3DGS), and the latter 145ms was measured using the same differentiable Monte-Carlo renderer as in training. Given that the textured mesh has been obtained in the former step, the rendering time can be significantly reduced by advanced rendering techniques (e.g. NIVIDIA RTX), making real-time rendering feasible in the future.

\begin{table}[t]
\caption{Quantitative Comparisons on \textit{SyntheticHuman++} dataset. Following \cite{xu2023relightable}, Normal degree and PSNR are computed only in the pixels with foreground mask activated, while SSIM and LPIPS are computed in the bounding box of the human region. The metric computations are slightly different to those in the main text. The best results and the second best are highlighted in \textbf{bold} and \underline{underlined} fonts, respectively.}
\label{tab:syn}
\scriptsize
\centering
\begin{tabular}{@{}lcccccccccc@{}}
\toprule
                                         & \textbf{Normal}    & \multicolumn{3}{c}{\textbf{Albedo}}                              & \multicolumn{3}{c}{\textbf{Relighting}}                          & \multicolumn{3}{c}{\textbf{Visibility}}                          \\ \midrule
                                         & Degree$\downarrow$ & PSNR$\uparrow$   & SSIM$\uparrow$   & LPIPS$\downarrow$ & PSNR$\uparrow$   & SSIM$\uparrow$   & LPIPS$\downarrow$ & PSNR$\uparrow$   & SSIM$\uparrow$   & LPIPS$\downarrow$ \\ \midrule
Ours                                 & $\underline{13.44}$            & $\mathbf{31.94}$ & $\mathbf{0.953}$ & $\mathbf{0.073}$  & $\mathbf{27.19}$ & $\mathbf{0.941}$ & $\mathbf{0.066}$  & $\mathbf{22.30}$ & $\mathbf{0.910}$ & $\mathbf{0.086}$  \\
Ours  (w/o normal)                   & $19.37$            & $\underline{31.08}$ & $\underline{0.942}$ & $\underline{0.072}$  & $\underline{26.92}$ & $\underline{0.939}$ & $\underline{0.068}$  & $\underline{20.36}$ & $\underline{0.891}$ & $\underline{0.098}$  \\
Xu \etal~\cite{xu2023relightable}        & $\mathbf{12.44}$   & $29.01$          & $0.933$          & $0.119$           & $22.69$          & $0.861$          & $0.206$           & $20.20$          & $0.848$          & $0.155$           \\
Relighting4D~\cite{chen2022relighting4d} & $29.38$            & $24.70$          & $0.885$          & $0.183$           & $22.13$          & $0.835$          & $0.237$           & $15.22$          & $0.763$          & $0.252$           \\
NeRFactor~\cite{zhang2021nerfactor}      & $34.29$            & $22.23$          & $0.817$          & $0.226$           & $21.04$          & $0.758$          & $0.313$           & $11.37$          & $0.581$          & $0.387$           \\ \bottomrule
\end{tabular}
\end{table}

\begin{figure}[t]
    \centering
    \includegraphics[width=\linewidth]{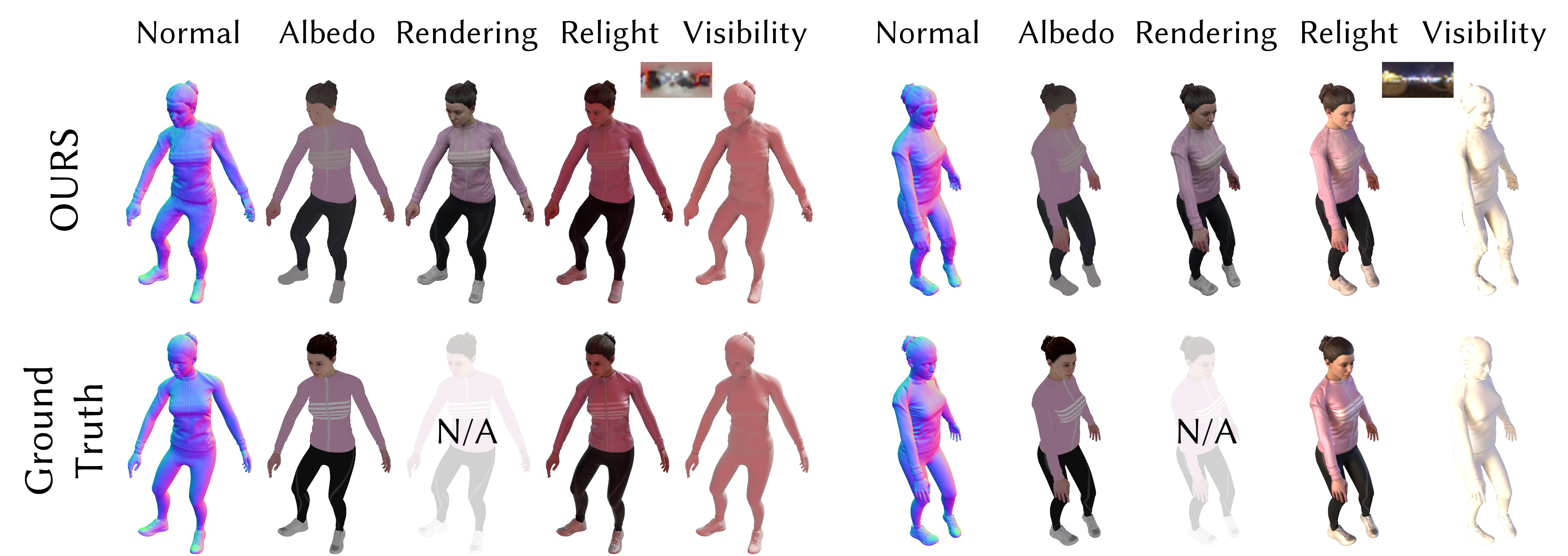}
    \caption{Visualizations of our method on character \textit{jody} in \textit{SyntheticHuman++} dataset.}
    \label{fig:syn}
\end{figure}

\section{More Validations on Inverse Rendering and Relighting}
\label{sec:more_exp}

\subsection{Evaluations on Synthetic Data}

For the task of intrinsic decomposition, we further evaluate our method on a synthetic dataset SyntheticHuman++~\cite{peng2021animatable} and compare with state-of-the-art methods~\cite{xu2023relightable,chen2022relighting4d,zhang2021nerfactor}. This dataset consists of $100$ frames $\times$ $20$ camera views, from which we use $10$ for training and the others for novel view testing. We use this original training/testing split in the dataset for evaluation, and follow the same protocol in \cite{xu2023relightable}. The results of \cite{xu2023relightable,chen2022relighting4d,zhang2021nerfactor} are borrowed from \cite{xu2023relightable}. Due to the scale amibiguity of inverse rendering problem, the metrics are computed after scale alignment~\cite{xu2023relightable,WangCVPR2024IntrinsicAvatar,lin2024relightable} on each color channel.     
As demonstrated in Table \ref{tab:syn}, our method significantly outperforms Relighting4D~\cite{chen2022relighting4d} across all inverse rendering metrics. Furthermore, benefiting from explicit mesh representation and more accurate rendering, our method also achieved notable improvement from Xu \etal~\cite{xu2023relightable} on the quality of novel light synthesis, despite limited enhancements in geometry reconstruction.
Example qualitative results of our method are shown in Figure \ref{fig:syn}.

We also evaluated the effectiveness of pseudo normal supervision on this synthetic dataset. Tab. \ref{tab:syn} presents the quantitative result without using estimated normals, denoted as \textit{Ours (w/o normal)}. In comparison with \textit{Ours}, it demonstrates that introducing geometric priors enhances the accuracy of geometry reconstruction, which subsequently improves image synthesis under novel lighting. We still achieved better performance than other SOTA methods in most metrics, except geometric error, underscoring the efficacy of our proposed method.

\begin{table}[t]
\scriptsize
\centering
\caption{Additional quantitative comparisons on reconstructions and novel pose synthesis. The better one is highlighted in \textbf{bold} fonts. }
\label{tab:avatar-2}
\begin{tabular}{@{}ccccccccc@{}}
\toprule
                & \multicolumn{4}{c}{Training Frame Reconstruction}                                           & \multicolumn{4}{c}{Novel Pose Synthesis}                                                    \\ \midrule
                & PSNR $\uparrow$       & SSIM $\uparrow$      & LPIPS $\downarrow$   & FID $\downarrow$      & PSNR $\uparrow$       & SSIM $\uparrow$      & LPIPS $\downarrow$   & FID $\downarrow$      \\ \midrule
AvatarReX       & $28.4015$             & $0.9607$             & $0.0579$             & $30.6214$             & $26.5853$             & $\mathbf{0.9497}$             & $\mathbf{0.0684}$ & $36.1911$             \\
Ours (w/o normal)      & $\mathbf{30.2084}$ & $\mathbf{0.9616}$ & $\mathbf{0.0564}$ & $\mathbf{24.7444}$ & $\mathbf{27.1589}$    & $0.9495$    & $0.0687$             & $\mathbf{31.2888}$ \\ \bottomrule
\end{tabular}
\end{table}

\begin{figure}[t]
    \centering
    \includegraphics[width=\linewidth]{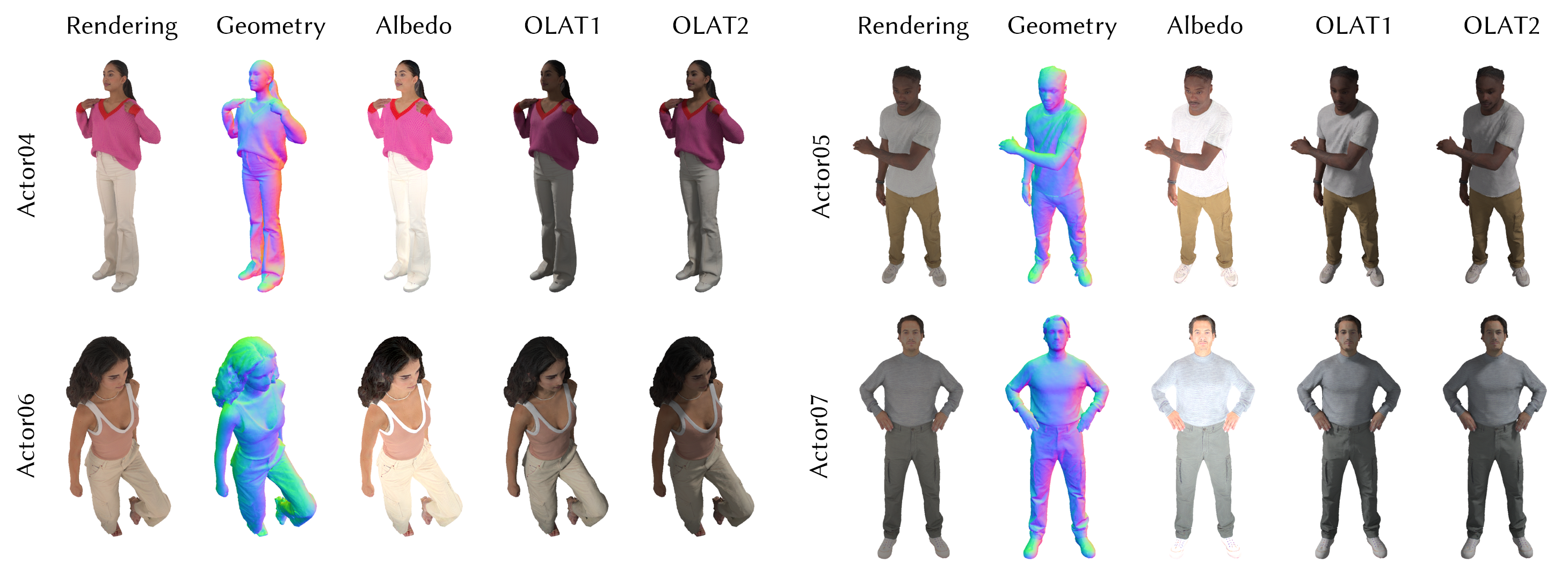}
    \caption{Visualizations of our learned avatars synthesized under OLAT environments.}
    \label{fig:olat}
\end{figure}

\begin{figure}[th]
    \centering
    \includegraphics[width=\linewidth]{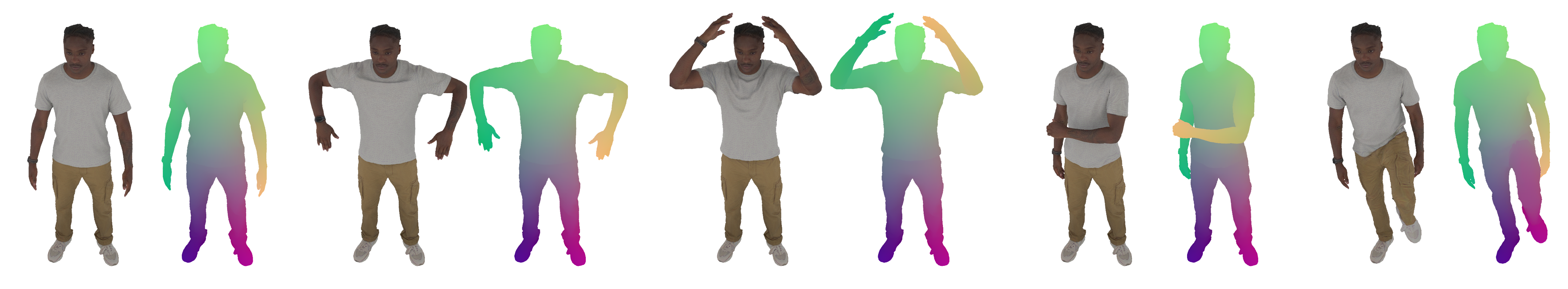}
    \caption{Visualizations of the correspondences of our learned avatar.}
    \label{fig:corr}
\end{figure}

\subsection{Additional Experiments}

    \textbf{Relighting under OLAT.} To further demonstrate the solved intrinsic properties and the relighting capability of our method, we relight our learned avatar using OLAT (One Light At a Time) environment maps and visualize them in Figure~\ref{fig:olat}. As shown in this figure, our method is able to realistically synthesize the shading effects of cloth wrinkles under different lighting directions. This experiment further proves that our method is able to recovery accurate geometry and albedo/material for dynamic humans. 

    \noindent \textbf{Comparisons without Normal Priors.} Considering our method employed additional priors from pseudo normal supervision, which may bias the comparison in the main text, we further report the quantitative result without using estimated normals in Tab. \ref{tab:avatar-2}. The evaluation is performed on AvatarReX dataset and the metrics are the same as we used in Tab. \ref{tab:avatar} in the main text.

    \noindent \textbf{Correspondences.} Another advantage of our method is that it naturally realizes surface tracking and establishes point-to-point correspondence among the whole performance sequence, which is typically difficult, if not impossible, in previous implicit representations. Figure \ref{fig:corr} shows the color-coded correspondences across different poses. The rendered colors on the right image of each sub-figure are defined as the corresponding normalized canonical coordinates of the ray-traced points. It demonstrates that our method learns reasonable mesh correspondences from images without explicit surface tracking.

\section{Potential Social Impacts}
\label{sec:social_impacts}

    Our method facilitates the automatic digital image creation of any specific human identity. However, this capability poses the risk to generate fake motion sequences that the individual has never performed. This issue should be carefully addressed before deployment.

\end{document}